\documentclass{article} 
\usepackage{iclr2027_conference,times}

\usepackage{amsmath,amsfonts,bm}

\newcommand{\newterm}[1]{{\bf #1}}

\def\Figref#1{Figure~\ref{#1}}

\def\Secref#1{Section~\ref{#1}}

\def\eqref#1{equation~\ref{#1}}
\def\Eqref#1{Equation~\ref{#1}}

\def\1{\bm{1}}
\newcommand{\train}{\mathcal{D}}

\def\eps{{\epsilon}}

\def\ry{{\textnormal{y}}}

\def\erva{{\textnormal{a}}}

\def\vmu{{\bm{\mu}}}
\def\vtheta{{\bm{\theta}}}
\def\va{{\bm{a}}}

\def\vf{{\bm{f}}}
\def\vg{{\bm{g}}}
\def\vh{{\bm{h}}}

\def\vm{{\bm{m}}}

\def\vq{{\bm{q}}}

\def\vt{{\bm{t}}}

\def\vx{{\bm{x}}}

\def\vz{{\bm{z}}}

\def\eva{{a}}

\def\mF{{\bm{F}}}

\def\mI{{\bm{I}}}

\DeclareMathAlphabet{\mathsfit}{\encodingdefault}{\sfdefault}{m}{sl}
\SetMathAlphabet{\mathsfit}{bold}{\encodingdefault}{\sfdefault}{bx}{n}

\def\gG{{\mathcal{G}}}

\def\gP{{\mathcal{P}}}

\def\gS{{\mathcal{S}}}

\def\gX{{\mathcal{X}}}

\def\sK{{\mathbb{K}}}

\def\sV{{\mathbb{V}}}

\def\sY{{\mathbb{Y}}}

\newcommand{\E}{\mathbb{E}}
\newcommand{\Ls}{\mathcal{L}}
\newcommand{\R}{\mathbb{R}}

\usepackage{hyperref}
\usepackage{url}
\usepackage{booktabs}       
\usepackage{caption}
\usepackage{graphicx}
\usepackage{multirow}
\usepackage{wrapfig}
\usepackage{tabularx}
\usepackage{array}
\usepackage{amsthm}
\usepackage{enumitem}
\usepackage{multicol}
\usepackage{colortbl}
\usepackage{makecell}

\title{Coverage-aware Semantic Representation Learning for Underrepresented Visual Concepts}

\author{%
  Sakib Ahammed$^{1}$\thanks{Corresponding author. \quad Email: {\tt\small SAKIB.AHAMMED@stu.mmu.ac.uk, \{X.Cui, W.Lu, X.Fan, b.cassidy, M.Yap\}@mmu.ac.uk, xuchen@hkmu.edu.hk}} \And  Xia Cui$^{1}$ \And Wenqi Lu$^{1}$ \And Xinqi Fan$^{1}$ \And Bill Cassidy$^{1}$ \And Xueli Chen$^{2}$ \And Moi Hoon Yap$^{1}$  \\ \\
  $^{1}$School of Computing and Mathematics, Manchester Metropolitan University, Manchester, UK \\
  $^{2}$School of Science and Technology, Hong Kong Metropolitan University, Hong Kong, China
}

\newcommand{\cc}{class--concept}
\newcommand{\ccpair}{\cc{} pair}
\newcommand{\ccpairs}{\cc{} pairs}
\newcommand{\SCI}{\textsc{SCI}}
\newcommand{\CCSS}{\textsc{CCSS}}
\newcommand{\SemCovNet}{\textsc{SemCovNet}}

\theoremstyle{plain} 
\newtheorem{definition}{Definition}[section]

\theoremstyle{remark}
\newtheorem{remark}{Remark}[section]

\iclrfinalcopy 
\begin{document}

\maketitle

\addtocontents{toc}{\protect\setcounter{tocdepth}{-1}}

\begin{abstract}
Modern vision models increasingly rely on rich semantic representations that extend beyond class labels to include descriptive concepts, attributes, and contextual cues. However, semantic concepts are not uniformly represented across classes: a concept may be frequent globally yet remain underrepresented within a class, resulting in low \cc{} coverage.
We formalize this \cc{} phenomenon as \textit{Semantic Coverage Imbalance (SCI)} and study its relationship with learned semantic representations.
We find that lower \cc{} coverage is consistently associated with weaker semantic representations.
To address this, we introduce \textit{Coverage-aware Semantic Representation Learning}, which shares more semantic structure for low coverage or uncertain relations while preserving more pair-specific structure for high coverage relations. We instantiate this principle in \SemCovNet{} with \textit{Coverage-Calibrated Semantic Sharing (CCSS)}, a coverage- and uncertainty-aware partial-pooling mechanism. Using a range of vision datasets spanning facial attributes and fine-grained recognition, as well as real-world medical datasets, we show that SCI is widespread, degrades representation quality, and can be mitigated through coverage-aware semantic representation learning.
Code is available at {\url{https://github.com/mmu-dermatology-research/SemCovNet}}
\end{abstract}

\section{Introduction} \label{sec:intro}

\begin{figure}[t]
    \centering
    \includegraphics[width=\linewidth]{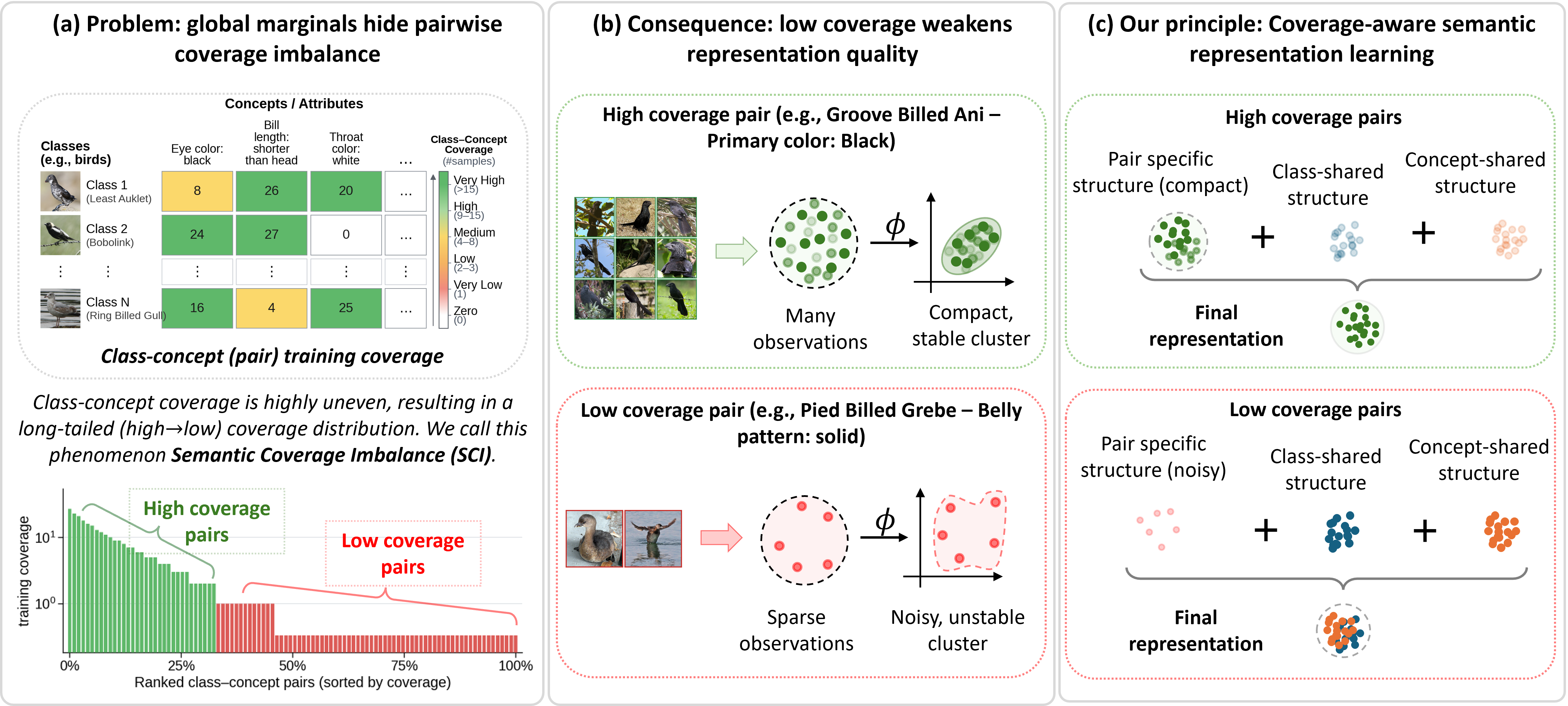}
     \caption{
    \textbf{Motivation and principle of coverage-aware semantic representation learning.}
        \textbf{(a)} Semantic Coverage Imbalance (\SCI{}) occurs when \ccpair{} coverage is highly uneven, even when class and concept marginals appear adequate.
        \textbf{(b)} We show that low coverage weakens semantic representations.
        \textbf{(c)} Coverage-aware semantic representation learning adapts more pair-specific structure for high coverage pairs and shares more structure for low coverage or uncertain pairs.
    }
    \label{fig:fig1-motivation}
\end{figure}

Semantic concepts provide an interpretable link between visual observations and task-level predictions, capturing class-relevant structure such as attributes, object parts, contextual cues, and clinically meaningful findings in medical imaging~\citep{koh2020cbm,zhang2025albm}. Concept-based models make such structure explicit and can support representations that are both discriminative and semantically meaningful~\citep{gao2026escaping}.
However, semantic concepts are not uniformly represented across classes \citep{zeng2024understanding,liu2025a,choi2026conceptscope}.
We use \newterm{semantic evidence} to denote the concept annotations or scores observed in training examples of a given class.
A concept may be common overall yet receive minimal coverage within a particular class, resulting in low 
coverage, as illustrated in \Figref{fig:fig1-motivation}(a).
This raises a fundamental question: \textit{how does the amount of semantic evidence for class and concept relationships affect their learned representation, and can semantic sharing be adapted to that evidence?}

We study this at the \ccpair{} level, since the same concept can have very different coverage across classes.
Prior work has demonstrated within-class attribute imbalance and class-dependent concept distributions
\citep{tang2022invariant,zhou2025class,choi2026conceptscope}, but it does not formalize the empirical coverage of each \ccpair{} or study how this coverage affects the learned semantic representation.
For example, an attribute may be common across bird species yet rare within one species. 
As \Figref{fig:fig1-motivation}(a) illustrates, class and concept marginals can therefore overlook a long-tailed distribution of \cc{} coverage.
We refer to this phenomenon as \newterm{Semantic Coverage Imbalance (SCI)}, uneven empirical coverage of \ccpairs{}, characterized by a long-tailed coverage distribution within a dataset.

SCI matters because low \cc{} coverage can make semantic representations less reliable.
As illustrated in \Figref{fig:fig1-motivation}(b), high coverage pairs are supported by many observations and result in compact representations, whereas low coverage pairs have sparse observations and can yield noisier representations. 
We examine this relationship in \Secref{subsec:rep-quality} 
and find that lower \cc{} coverage is associated with weaker semantic representations.
This is especially important in medical imaging, where diagnosis depends on clinically meaningful visual findings \citep{yang2024textbook,gao2024aligning,huy2025interactive}.
In Derm7pt, rare diagnoses and rare findings create clinically important low coverage \ccpairs{} despite expert-defined diagnostic criteria for the 7-point melanoma checklist~\citep{Kawahara2018-7pt}.

Long-tailed methods mainly rebalance classes \citep{Kang2020Decoupling,menon2021longtail,luo2026long}, while generalized long-tail learning considers within-class attribute imbalance \citep{tang2022invariant,zhou2025class}. 
Group-robust methods target underperforming subpopulations \citep{sagawa2020groupdro,liu2021jtt,park2026spurious}, and concept-based methods improve grounding, class-specific structure, or semantic representations \citep{zhang2025albm,gao2026escaping,choi2026conceptscope}.
However, existing methods generally optimize class frequencies, predefined groups, or attribute distributions rather than using \cc{} relations to show how they learn semantic representations. 
We discuss related work in more detail in Appendix~\ref{app:related_work}.

\paragraph{Contributions.}
To address this limitation, we identify and formalize \newterm{Semantic Coverage Imbalance (\SCI{})}.
We introduce \newterm{Coverage-aware Semantic Representation Learning}, illustrated in \Figref{fig:fig1-motivation}\textbf{(c)}, which shares more semantic structure for low coverage or uncertain relations while preserving pair-specific structure for high coverage relations. 
We instantiate this principle in \newterm{\SemCovNet{}}, a coverage-aware semantic representation learning framework.
Its \newterm{Coverage-Calibrated Semantic Sharing (\CCSS{})} mechanism adapts sharing using pair support and representation uncertainty.
Across CelebA, CUB, Waterbirds, Derm7pt, and MILK10k, spanning binary, categorical, expert-defined, and soft semantic evidence, we find that \SCI{} is widespread and that lower \cc{} coverage is associated with weaker semantic representations.
Controlled coverage interventions further show that reducing pair-specific coverage degrades representation quality, while coverage-aware learning mitigates this effect.
Overall, our results suggest that semantic supervision depends not only on which concepts are annotated, but also on \textit{how} their coverage is distributed across \cc{} relations.
We next formalize \SCI{} and its representation-level consequence.

\section{Formalizing Semantic Coverage Imbalance in Visual Learning}
\label{sec:sci} 

\begin{wrapfigure}[25]{r}[0pt]{0.44\linewidth} 
    \vspace{-12pt} 
    \centering
    \includegraphics[width=0.95\linewidth, trim={0pt 2.5pt 0pt 5pt}, clip]{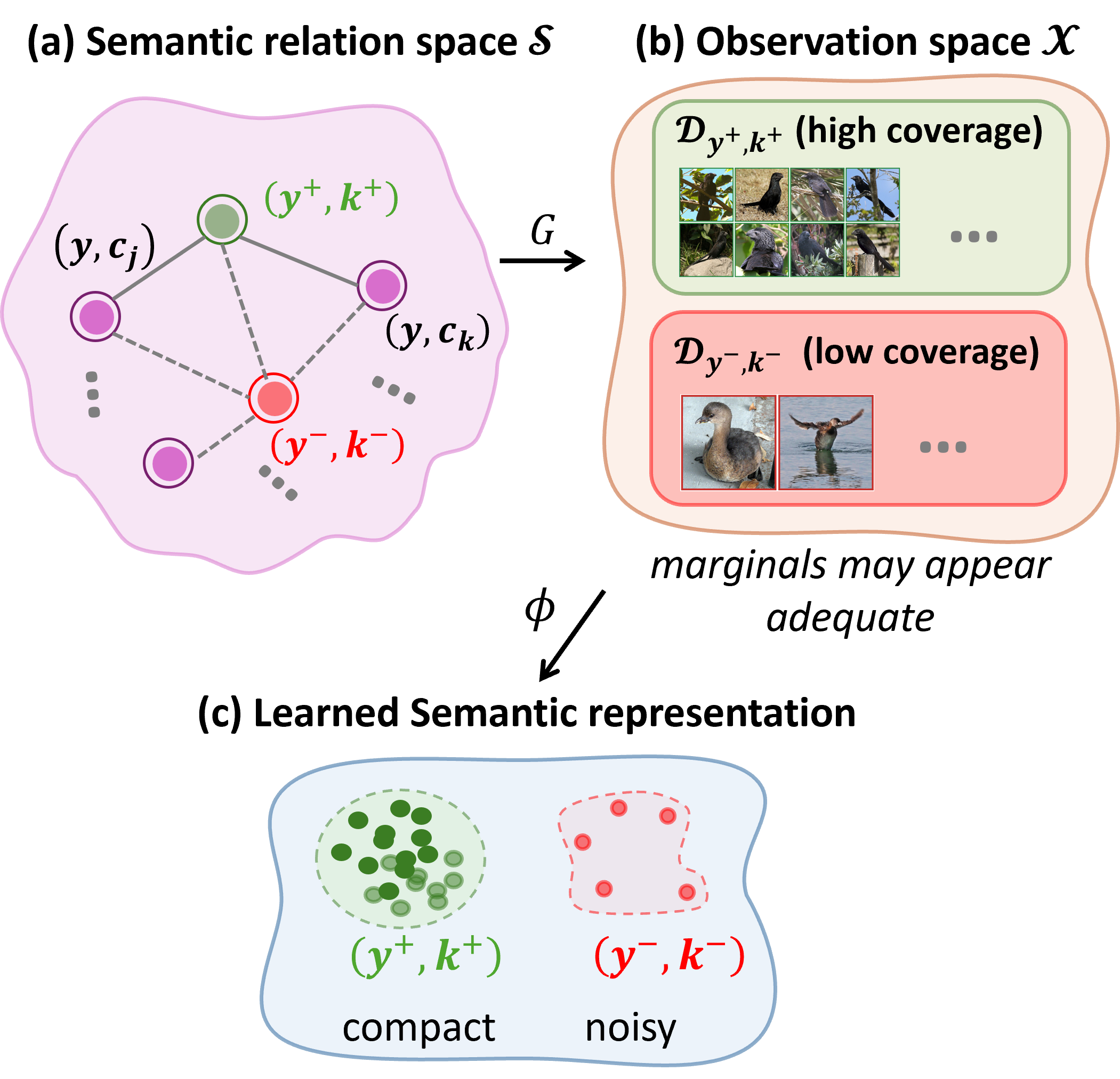} 
    \caption{
    \textbf{Intuitive view of Semantic Coverage Imbalance (\SCI{}).}
    \textbf{(a)} Each \ccpair{} $(y,k)$ defines a relation in the semantic space $\gS$.
    \textbf{(b)} In the observation space $\gX$, the training data grounds each semantic relation as $\gG:(y,k)\mapsto\train_{y,k}$, producing high- and low coverage pairs.
    \textbf{(c)} We hypothesize that low coverage leads to weaker learned semantic representations. We refer to the phenomenon of uneven \ccpair{} coverage as \SCI{}.
    }
    \label{fig:fig2-sci}
\end{wrapfigure}

We now formalize \newterm{Semantic Coverage Imbalance (\SCI{})}. \Figref{fig:fig2-sci} illustrates the intuitive view of SCI: how latent semantic relations are represented in observed training evidence and the resulting learned representations.

\subsection{Problem Setup}
\label{subsec:sci_setup}

Let $\gX$ denote the observation space, $\sY := \{1,\ldots,T\}$ the set of
class labels, and $\sK := \{1,\ldots,K\}$ the set of semantic concepts
(e.g., attributes, parts, or clinical findings). The training set is
\(
    \train
        =
    \left\{
        (\vx_i,y_i,\va_i)
    \right\}_{i=1}^{n},
\)
$\va_i\in[0,1]^K,$ where $\eva_{i,k}$ denotes the evidence for concept $k$ in image $\vx_i$. This definition is used for binary attributes, categorical attributes expanded into atomic concepts, expert
annotations, and soft semantic evidence.

Each \newterm{\ccpair{} $(y,k)$} defines a relation in a semantic
relation space $\gS$, which the training data grounds in a set of
observed examples by mapping $\gG:(y,k)\mapsto\train_{y,k}$, where
$\train_{y,k}$ contains the observations that provide evidence for
$(y,k)$.
For binary, multi-level, or categorical annotations, the observations
that support pair $(y,k)$ are
\(
    \train_{y,k}
        :=
    \left\{
        \vx_i :
        y_i=y,\;
        a_{i,k}=1
    \right\}.
\)
For soft semantic annotations, pair support is represented by the
corresponding evidence mass.

\subsection{Class--Concept Coverage}
\label{subsec:sci_coverage}

We measure the available semantic evidence through class-conditioned concept coverage.

\begin{definition}[\cc{} coverage]
\label{def:cccov}
Let $n_y := \sum_{i=1}^{n}\1\{y_i=y\}$. The empirical coverage of class $y$ and concept $k$ is
\begin{equation}
    C_{y,k} := \E[\erva_k\mid\ry=y]
    =
    \frac{1}{n_y}
    \sum_{i:y_i=y} a_{i,k}.
    \label{eq:class_concept_coverage}
\end{equation}
\end{definition}
For binary concepts, $C_{y,k}$ is the within-class prevalence of concept $k$; for soft evidence, it is the mean evidence mass within class $y$. We use \textit{coverage} to denote the amount of observed support available for learning relation $(y,k)$. We also define the effective support as the cardinality of $\train_{y,k}$ as
\begin{equation}
    n^{\mathrm{eff}}_{y,k}
    :=
    |\train_{y,k}|
    =
    \sum_{i:y_i=y} a_{i,k},
    \label{eq:n_effective_support}
\end{equation}
the total observed evidence contributing to pair $(y,k)$. We refer to a pair with large $C_{y,k}$ (or $n^{\mathrm{eff}}_{y,k}$) as \textit{high coverage} (well-supported), and a pair with small $C_{y,k}$ as \textit{low coverage} (undercovered).

\textbf{Property.} 
\cc{} coverage is pair-specific and cannot be determined from class or concept marginals alone. Since
\(
    \E[\erva_k]
    =
    \sum_{y=1}^{T}
    P(\ry=y)\,
    C_{y,k},
\)
the global prevalence of concept $k$ is only a class-frequency-weighted average of its class-conditioned coverage.
This pairwise view is what separates semantic coverage from standard class or concept frequency (\Figref{fig:fig2-sci}).

Consequently, coverage must be considered over the full set of \ccpairs{}. When ranked by $C_{y,k}$, these pairs can exhibit a substantial low coverage tail, as illustrated in \Figref{fig:fig1-motivation}(a). We formalize this uneven \ccpair{} coverage distribution as \SCI{} in \Secref{subsec:sci_definition}.


\subsection{Semantic Coverage Imbalance}
\label{subsec:sci_definition}


\begin{definition}[Semantic Coverage Imbalance]
\label{def:sci}
Let
\(
    \gP
    :=
    \{(y,k)\in\sY\times\sK : n_y>0\}
\)
denote the valid \ccpairs{} in the training set, with empirical coverage $C_{y,k}$ defined in \Eqref{eq:class_concept_coverage}.
We define \newterm{Semantic Coverage Imbalance (\SCI{})} as the uneven distribution of empirical coverage
\(
    \{C_{y,k}\}_{(y,k)\in\gP}
\)
across \ccpairs{} within a dataset.

\SCI{} is a pair-level property and is not determined by class or concept marginals alone.
In particular, a \ccpair{} may have low coverage even when its class is common, and its concept is frequently observed overall.
Thus, class and concept marginals can appear well represented overall while their specific \cc{} relation has low coverage.
\end{definition}

\begin{remark}[Long-tailed structure of \cc{} coverage]
\label{rem:sci-long-tail}
Ranking the coverage values as 
$C_{(1)} \ge C_{(2)} \ge \cdots \ge C_{(N)}$,
where $N=|\gP|$, provides the \ccpair{} analogue of the ranked class-frequency curve commonly used to characterize long-tailed label distributions \citep{cui2019class}.
In practice, \SCI{} appears as a long low coverage tail containing many \ccpairs{}, rather than as a single rare relation. Appendix~\ref{app:sci-definition} provides a formal SCI characterization.
\end{remark}

\paragraph{Representation hypothesis.}
\label{subsec:representation-hypothesis}
\Figref{fig:fig2-sci}(c) illustrates the representation hypothesis underlying our study.
Let $\phi:\gX\rightarrow\R^D$ denote a learned visual representation, and let $q_{y,k}(\phi)$ measure the representation quality of \ccpair{} $(y,k)$.
We hypothesize that
\begin{equation}
    C_{y,k}\downarrow
    \;\Longrightarrow\;
    \E[q_{y,k}(\phi)]\downarrow,
    \label{eq:rep_hypothesis}
\end{equation}
i.e., lower \cc{} coverage corresponds to lower expected representation quality.
Importantly, \SCI{} is a property of the training distribution and is defined independently of the learned model.
Whether lower coverage is associated with weaker semantic representations is therefore an empirical question, which we examine in \Secref{subsec:rep-quality} and further test through controlled coverage interventions in \Secref{subsec:controlled-intervention}.
\section{Method: Coverage-Aware Semantic Representation Learning}
\label{sec:method}

\begin{figure}[t]
    \centering
    \includegraphics[width=\textwidth]{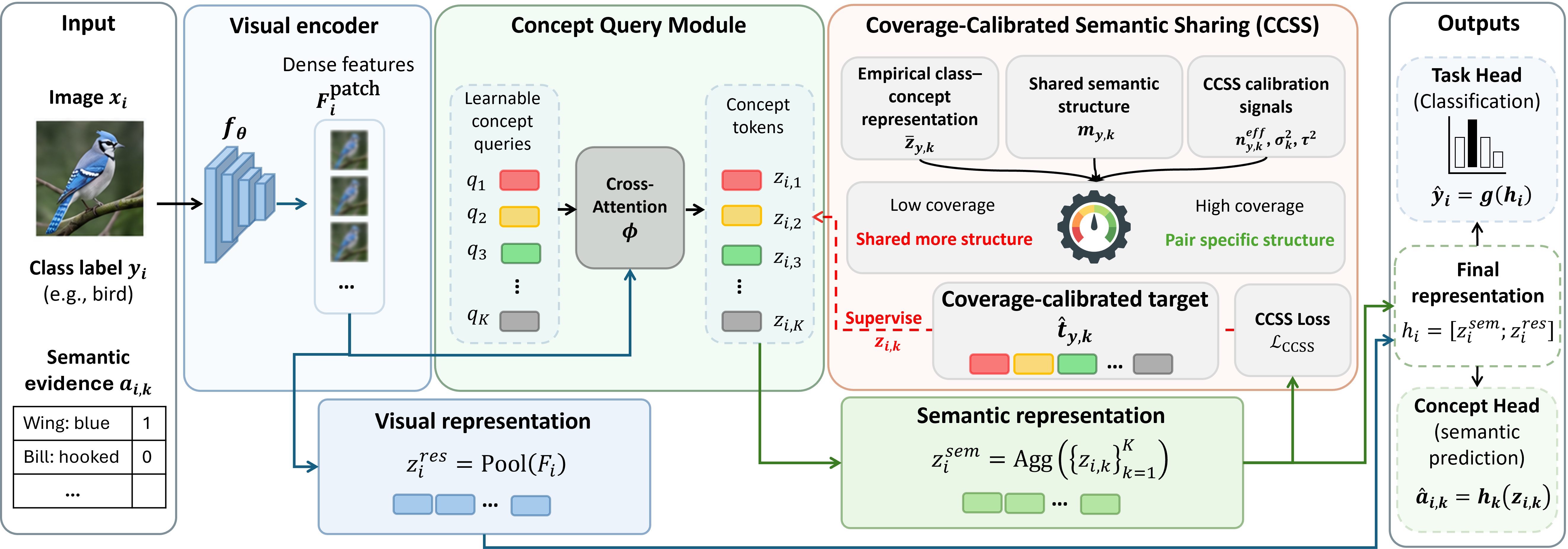}
    \caption{
    \textbf{\SemCovNet{} for coverage-aware semantic representation learning.} 
    A Concept Query Module maps dense features to image-driven concept tokens.
    For each \ccpair{}, \CCSS{} combines the empirical \cc{} representation,
    a shared semantic structure, and calibration signals.
    low coverage or uncertain pairs shrink toward the shared structure, while high coverage pairs retain more pair-specific structure.
    The resulting target supervises the concept tokens, and a residual pathway preserves complementary visual information for task prediction.
    }
    \label{fig:fig3-semcovnet}
\end{figure}

\Secref{sec:sci} formalizes \SCI{} as uneven \ccpair{} coverage and motivates the hypothesis that low coverage can weaken semantic representations.
We therefore instantiate our proposed \newterm{Coverage-aware Semantic Representation Learning} principle in \textbf{\SemCovNet{}}, a coverage-aware semantic representation learning framework:high coverage pairs retain more pair-specific structure, while low coverage or uncertain pairs share more semantic structure.
As illustrated in \Figref{fig:fig3-semcovnet}, a Concept Query Module maps dense image features to visual concept tokens for each semantic concept.
\newterm{Coverage-Calibrated Semantic Sharing (\CCSS{})} then determines how strongly each \ccpair{} relies on shared semantic structure versus its observed pair-specific representation.

\subsection{Learning Visual Concepts}
\label{subsec:concept_tokens}

Given image $\vx_{i}$, a visual encoder $f_{\vtheta}$ produces dense patch
features
\(
    \mF_{i}^{\mathrm{patch}} = f_{\vtheta}(\vx_{i}) = \{\vf_{i,p}\}_{p=1}^{P}.
\)
In our implementation, we initialize $f_{\vtheta}$ from DINOv3\footnote{We use DINOv3 as the visual encoder: its self-supervised
pretraining on large-scale unlabeled data yields structured representations that transfer across tasks (classification, segmentation, depth, tracking) and domains (natural, medical, satellite) without text supervision~\citep{simeoni2025dinov3}}. 
For each semantic concept $k$, we use a learnable concept query embedding $\vq_k\in\R^D$. 

A Concept Query Module uses cross-attention to extract the visual realization of each concept:
\begin{equation}
    \vz_{i,k}
    =
    \Phi
    \left(
        \vq_k,
        \mF_{i}^{\mathrm{patch}}
    \right).
    \label{eq:concept_query_module}
\end{equation}
Here, $\Phi$ denotes the cross-attention operator and $\vz_{i,k}\in\R^D$ is the resulting visual concept token.

The semantic evidence $\eva_{i,k}$ is \textit{not} an input to \Eqref{eq:concept_query_module}. The concept token $\vz_{i,k}$ is inferred from the image alone, while $\eva_{i,k}$ is used only as training supervision: $\hat{\eva}_{i,k} = h_{k}(\vz_{i,k})$, giving the semantic supervision loss $\Ls_{\mathrm{concept}}$ in \Eqref{eq:task_concept_losses}.

\subsection{Pair-Specific and Shared Semantic Structure}
\label{subsec:shared_structure}

We estimate the observed \cc{} pair representation by aggregating the concept tokens from \Eqref{eq:concept_query_module} using the available semantic evidence:
\begin{equation}
    \bar{\vz}_{y,k}
    =
    \frac{
        \sum_{i:y_i=y}
        \eva_{i,k}\vz_{i,k}
    }{
        n^{\mathrm{eff}}_{y,k}+\eps
    }.
    \label{eq:empirical_pair_representation}
\end{equation}
The prototype $\bar{\vz}_{y,k}$ is a stop-gradient statistic computed only from the training split for each \ccpair{}. 

We define a shared semantic structure as:
\begin{equation}
    \vm_{y,k}
    =
    \underbrace{\vmu_0}_{\text{global structure}}
    +
    \underbrace{\tilde{\vmu}_y}_{\text{class-shared structure}}
    +
    \underbrace{\tilde{\vmu}_k}_{\text{concept-shared structure}},
    \label{eq:shared_semantic_prior}
\end{equation}
where $\vmu_0\in\R^D$ is a global learnable vector shared by classes and concepts.
$\vmu_y$ is a learnable vector for class $y$, shared by every concept in that class; $\vmu_k$ is a learnable vector for concept $k$, shared by every class it appears in. To make this decomposition identifiable, we use
\(
    \tilde{\vmu}_y
    =
    \vmu_y
    -
    \frac{1}{C}\sum_{y'=1}^{C}\vmu_{y'},
\)
and
\(
    \tilde{\vmu}_k
    =
    \vmu_k
    -
    \frac{1}{K}\sum_{k'=1}^{K}\vmu_{k'},
    \label{eq:prior_centering}
\)
so that each term in \Eqref{eq:shared_semantic_prior} captures a distinct source of structure, with no overlap between them.
Details are provided in Appendix~\ref{app:ccss_gradient}.

\subsection{Coverage-Calibrated Semantic Sharing (CCSS)}
\label{subsec:ccss}

The reliability of the aggregated concept token $\bar{\vz}_{y,k}$ depends on how much evidence supports the \cc{} pair and how visually variable the concept is. 
Therefore, we estimate the uncertainty of each concept $k$ as the variance of each concept token:
\begin{equation}
    \sigma_k^2
    =
    \frac{
        \sum_{y=1}^{C}\sum_{i:y_i=y}
        \eva_{i,k}
        \left\|\vz_{i,k}-\bar{\vz}_{y,k}\right\|_2^2
    }{
        D\sum_{y=1}^{C} n^{\mathrm{eff}}_{y,k}+\eps
    }.
    \label{eq:concept_uncertainty}
\end{equation}

We also estimate the uncertainty of the shared semantic structure defined in \Eqref{eq:shared_semantic_prior} as a global between-pair variance of how much pairs vary around their shared prior, net of sampling noise. Let $\sV$ be the set of valid \ccpairs{} with $n^{\mathrm{eff}}_{y,k}>0$:
\begin{equation}
    \tau^2
    =
    \left[
        \frac{1}{|\sV|}
        \sum_{(y,k)\in\sV}
        \left(
            \frac{\left\|\bar{\vz}_{y,k}-\vm_{y,k}\right\|_2^2}{D}
            -
            \frac{\sigma_k^2}{n^{\mathrm{eff}}_{y,k}}
        \right)
    \right]_{+}.
    \label{eq:tau2_estimator}
\end{equation}
The subtracted term removes the sampling noise expected purely from
finite support; $[\cdot]_+$ floors the result at zero, since a variance
cannot be negative.

CCSS combines $\sigma_k^2$, $\tau^2$, and effective support
$n^{\mathrm{eff}}_{y,k}$ (\Eqref{eq:n_effective_support}) into a \textit{pair-specificity weight}:
\begin{equation}
    \alpha_{y,k}
    =
    \frac{
        n^{\mathrm{eff}}_{y,k}\tau^2
    }{
        n^{\mathrm{eff}}_{y,k}\tau^2+\sigma_k^2
    }
    \in[0,1],
    \label{eq:ccss_weight}
\end{equation}
where $\alpha_{y,k}=0$ when $n^{\mathrm{eff}}_{y,k}=0$: a pair with no observed evidence relies entirely on the shared structure. The resulting coverage-calibrated target is
\begin{equation}
    \hat{\vt}_{y,k}
    =
    (1-\alpha_{y,k})\vm_{y,k}
    +
    \alpha_{y,k}\bar{\vz}_{y,k}.
    \label{eq:ccss_target}
\end{equation}

\CCSS{} therefore depends on shared structure when evidence is low coverage or uncertain, and increasingly allows class-concept specialization as stable evidence increases.

\subsection{Coverage-Aware Semantic Representation Learning Objective}
\label{subsec:ccss_learning}

We align each visual concept token with its coverage-calibrated target using semantic evidence weighted cosine loss:
\begin{equation}
    \Ls_{\mathrm{CCSS}}
    =
    \frac{
        \sum_{i=1}^{B}\sum_{k=1}^{K}
        \eva_{i,k}
        \left[
            1-
            \cos\!\left(
                \vz_{i,k},
                \hat{\vt}_{y_i,k}
            \right)
        \right]
    }{
        \sum_{i=1}^{B}\sum_{k=1}^{K}\eva_{i,k}
        +\eps
    }.
    \label{eq:ccss_loss}
\end{equation}
The target $\hat{\vt}_{y,k}$ in \Eqref{eq:ccss_target} is formed in the unnormalized latent space; we apply normalization only within the cosine similarity. Thus, low-support (low coverage) or uncertain pairs are regularized more strongly toward shared semantic structure, while well-supported (high coverage) pairs retain more pair-specific structure.

The annotated concept vocabulary may not capture all visual information.
We therefore retain a residual visual pathway,
\(
    \vz_i^{\mathrm{res}}
    =
    \operatorname{Pool}(\mF_i)
\),
and aggregate the concept tokens as
\(
    \vz_i^{\mathrm{sem}}
    =
    \operatorname{Agg}(\{\vz_{i,k}\}_{k=1}^{K})
\).
The final task representation is
\(
    \vh_i
    =
    \left[
        \vz_i^{\mathrm{sem}};
        \vz_i^{\mathrm{res}}
    \right].
\)

We optimize the task and semantic supervision objectives as
\begin{equation}
    \Ls_{\mathrm{task}}
    =
    \frac{1}{B}
    \sum_{i=1}^{B}
    \operatorname{CE}\!\left(
        g(\vh_i),\,y_i
    \right),
    \qquad
    \Ls_{\mathrm{concept}}
    =
    \frac{1}{BK}
    \sum_{i=1}^{B}
    \sum_{k=1}^{K}
    \ell_{\mathrm{sem}}\!\left(
        \hat{\eva}_{i,k},
        \eva_{i,k}
    \right),
    \label{eq:task_concept_losses}
\end{equation}
where $g$ is the task classifier,
$\hat{\eva}_{i,k}=h_k(\vz_{i,k})$ is the semantic prediction from
\Secref{subsec:concept_tokens}, and
$\ell_{\mathrm{sem}}$ denotes the dataset-specific semantic supervision loss (Appendix~\ref{app:implementation-details}).

The final objective is
\begin{equation}
    \Ls
    =
    \Ls_{\mathrm{task}}
    +
    \lambda_{\mathrm{concept}}\Ls_{\mathrm{concept}}
    +
    \lambda_{\mathrm{CCSS}}\Ls_{\mathrm{CCSS}},
    \label{eq:semcovnet_objective}
\end{equation}
where $\lambda_{\mathrm{concept}}$ and $\lambda_{\mathrm{CCSS}}$ are fixed scalar weights for the concept supervision and coverage-aware alignment losses.
We first warm up the Concept Query Module and prediction heads with $\Ls_{\mathrm{task}}+\lambda_{\mathrm{concept}}\Ls_{\mathrm{concept}}$, then compute the training-set CCSS statistics and activate $\Ls_{\mathrm{CCSS}}$ while adapting the selected DINOv3 blocks. $\bar{\vz}_{y,k}$, $\sigma_k^2$, and $\tau^2$ remain stop-gradient quantities throughout. 

Semantic evidence and CCSS statistics are used only during training.
Therefore, at inference, only the image is required:
\begin{equation}
    \vx_i
    \xrightarrow{f_{\vtheta}}
    \mF_i
    \longrightarrow
    \{\vz_{i,k}\}_{k=1}^{K},
    \quad
    \begin{cases}
        \vz_i^{\mathrm{sem}}
        =
        \operatorname{Agg}(\{\vz_{i,k}\}_{k=1}^{K}),
        \\[2pt]
        \vz_i^{\mathrm{res}}
        =
        \operatorname{Pool}(\mF_i),
    \end{cases}
    \longrightarrow
    \hat{y}_i = g(\vh_i),
    \quad
    \hat{\eva}_{i,k}
    =
    h_k(\vz_{i,k}).
    \label{eq:inference}
\end{equation}

\section{Results and Discussion}
\label{sec:results}

\subsection{Experimental Setup.}
We evaluate the proposed coverage-aware semantic representation learning on CelebA, CUB, Waterbirds, Derm7pt, and MILK10k. We use the available semantic evidence in each dataset, including binary attributes, expert-defined clinical concepts, and continuous soft semantic scores. 
We use DINOv3 as the visual encoder and conduct observational and controlled analyses, as well as comparisons with concept-learning, imbalance-aware, and robustness baselines under identical training protocols.
For analysis, we rank \ccpairs{} by training coverage $C_{y,k}$ and use the 25th and 75th percentiles to define Tail ($Q_{0\text{--}25\%}$), Middle ($Q_{25\text{--}75\%}$), and Head ($Q_{75\text{--}100\%}$) groups. \textbf{Tail/Middle/Head pairs are used only for reporting and visualization; \SemCovNet{} \textit{does not} use them as training thresholds.}
Appendix~\ref{sec:exp-setup} provides full details of the dataset, model, baselines, training, and evaluation protocol.

\subsection{SCI is Widespread Across Semantic Visual Datasets}
\label{sec:results-sci}

\begin{wrapfigure}{r}[0pt]{0.50\linewidth}
    \centering
    \includegraphics[width=\linewidth]{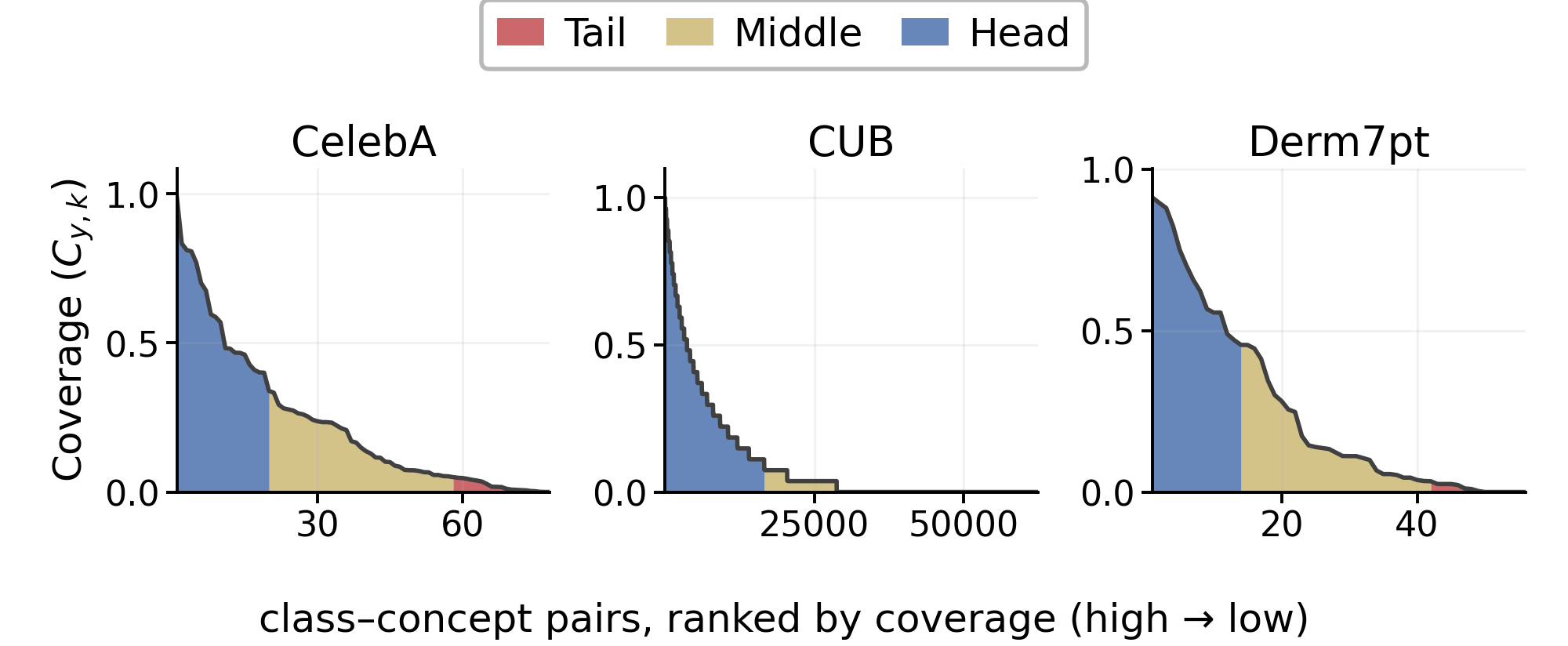}
    \caption{
    \textbf{\SCI{} across visual domains.}
    Ranked training-set \cc{} coverage $C_{y,k}$ for CelebA, CUB, and Derm7pt. 
    Coverage is highly uneven: 26.9\%, 67.2\%, and 33.9\% of \ccpairs{} have $C_{y,k}<0.05$ in CelebA, CUB, and Derm7pt, respectively.
    }
    \label{fig:fig4a-sci-distribution}
\end{wrapfigure}

We first investigate whether \SCI{} occurs in real training data. For each \ccpair{} $(y,k)$, we compute training coverage $C_{y,k}$. \Figref{fig:fig4a-sci-distribution} shows strongly uneven coverage across CelebA, CUB, and Derm7pt. The fraction of \ccpairs{} with $C_{y,k}<0.05$ is 26.9\%, 67.2\%, and 33.9\%, respectively; in CUB, 53.9\% of pairs have zero training coverage. These results show that low coverage \ccpairs{} are common across distinct visual domains.

Appendix~\ref{app:sci-characterization} provides a detailed characterization of \SCI{} across all five datasets, including Waterbirds and MILK10k. Appendix~\ref{app:sci-characterization} further shows that class frequency alone does not explain this structure: class-balanced resampling largely preserves the coverage ordering, and permutation tests show substantial across-class variation in concept coverage. We next examine whether lower coverage is associated with weaker semantic representations.

\begin{figure}[h]
    \centering
    \includegraphics[width=\textwidth]{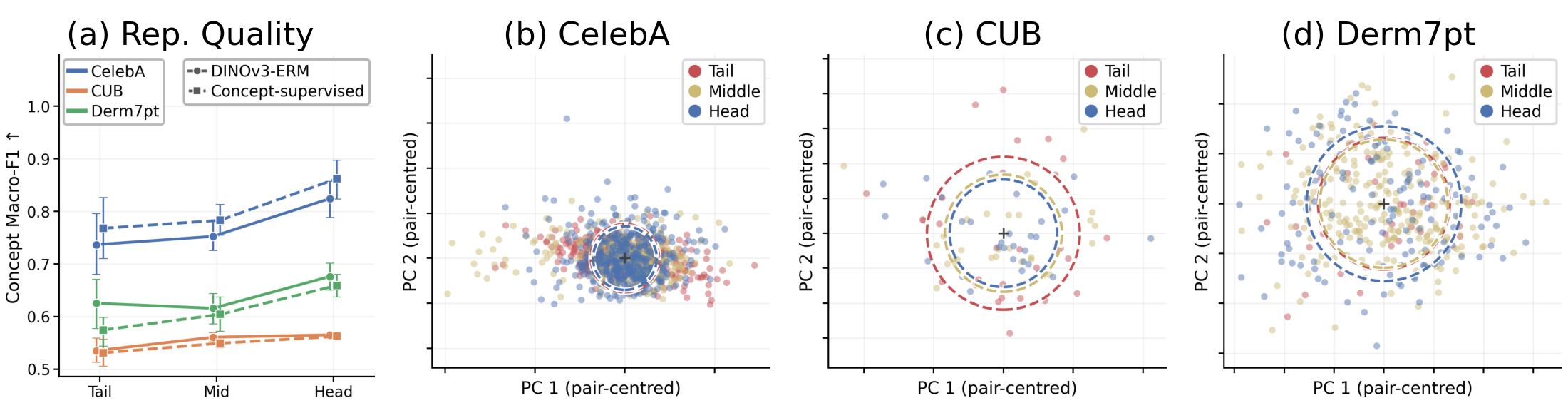}
    \caption{
    \textbf{Lower coverage is associated with weaker semantic representations.}
    \textbf{(a)} Frozen linear-probe class-conditioned concept Macro-F1 from the learned concept tokens across 
    Tail, Middle, and Head \ccpairs{} for DINOv3--ERM and concept supervised learning. Error bars show 95\% CIs over \ccpairs{}.
    \textbf{(b--d)} Concept supervised representations projected using a common PCA basis fitted separately for each dataset. Dispersion varies across coverage groups and datasets; $+$ denotes \cc{} centroids.
    }
    \label{fig:fig4b-coverage-representation}
\end{figure}

\subsection{Lower Coverage Is Associated with Weaker Semantic Representations}
\label{subsec:rep-quality}

\Figref{fig:fig4b-coverage-representation}(a) shows a clear coverage-dependent representation gap. Using the same frozen linear probe, DINOv3--ERM exhibits Head--Tail class-conditioned concept Macro-F1 gaps of $0.087$, $0.030$, and $0.051$ on CelebA, CUB, and Derm7pt, respectively. Concept supervision does not remove this gap: the corresponding gaps remain $0.094$, $0.031$, and $0.084$. Thus, semantic supervision alone does not eliminate the representation disparity between low- and high coverage \ccpairs{}. 
These results support our representation hypothesis in \Secref{subsec:sci_definition}.

The representation geometry provides complementary evidence. \Figref{fig:fig4b-coverage-representation}(b--d) shows greater Tail dispersion than Head dispersion on CelebA and CUB, decreasing from $7.12$ to $6.45$ and from $10.60$ to $8.77$, respectively. Derm7pt retains the frozen-probe coverage gap, but its geometric trend is less consistent; we therefore treat PCA geometry as qualitative supporting evidence. Overall, lower \cc{} coverage is associated with weaker semantic representations, motivating coverage-aware learning that shares more semantic structure for low coverage or uncertain pairs while preserving pair-specific structure for high coverage pairs.

\subsection{CCSS Improves low coverage Representations}
\label{sec:ccss-results}



We next test whether the proposed coverage-aware learning principle mitigates this gap. We test this in two steps. First, we isolate the effect of our proposed \newterm{Coverage-Calibrated Semantic Sharing (\CCSS{})} using a controlled model ladder in which matched variants differ only by adding \CCSS{}. Second, we compare \SemCovNet{} with existing representation-, imbalance-, and concept-learning SOTA methods.

\begin{table}[t]
\centering
    \caption{
    \textbf{Controlled CCSS model ladder on CUB and CelebA.}
    The ladder isolates the effect of CCSS under frozen and LoRA-adapted DINOv3 encoders (last three layers).
    Concept denotes standard semantic supervision without CCSS.
    Tail reports class-conditioned concept Macro-F1 and $\Delta_{\mathrm{H-T}}$ the Head--Tail representation gap. Highlighted rows show the paired Tail gain from adding CCSS under the same
    adaptation setting, with 95\% CIs.
    }
    \label{tab:ccss-ladder}

    \resizebox{\linewidth}{!}
    {
    \begin{tabular}{lccl|ccc|ccc}
    \toprule
    \multirow{2}{*}{Variant} & \multirow{2}{*}{\makecell{Concept\\supervised}} & \multirow{2}{*}{ CCSS} & \multirow{2}{*}{Adaptation} & \multicolumn{3}{c|}{CUB} & \multicolumn{3}{c}{CelebA} \\
    \cmidrule{5-7} \cmidrule{8-10}
    ~ & ~ & ~ & ~ &
    Task$\uparrow$ & Tail$\uparrow$ & $\Delta_{\mathrm{H-T}}$$\downarrow$ 
    & Task$\uparrow$ & Tail$\uparrow$ & $\Delta_{\mathrm{H-T}}$$\downarrow$ \\
    \midrule
    ERM &  $\times$  &  $\times$  & Frozen &
    0.886{\scriptsize$\pm$0.006}  & N/A &  N/A
    & 0.953{\scriptsize$\pm$0.001}  & N/A &  N/A \\
    
    \midrule
    
    Concept &  $\checkmark$ &  $\times$  &  Frozen & 
    0.881{\scriptsize$\pm$0.007} & 0.003{\scriptsize$\pm$0.002} & 0.171{\scriptsize$\pm$0.032} 
    & 0.954{\scriptsize$\pm$0.004} & 0.441{\scriptsize$\pm$0.102} & 0.426{\scriptsize$\pm$0.091} \\
    
    Concept$+$CCSS (Ours) &  $\checkmark$ &  $\checkmark$ &  Frozen & 
    0.888{\scriptsize$\pm$0.002} & 0.005{\scriptsize$\pm$0.001} & 0.145{\scriptsize$\pm$0.011} 
    & 0.956{\scriptsize$\pm$0.002} & 0.463{\scriptsize$\pm$0.006} & 0.406{\scriptsize$\pm$0.008} \\
    
    \rowcolor{purple!10}
    \multicolumn{4}{l|}{
    \textbf{$\Delta_{\mathrm{Tail}}$
    (Concept+CCSS $-$ Concept), Frozen [95\% CI]}
    } 
    & \multicolumn{3}{c|}{\textbf{0.002 \scriptsize[0.001, 0.002]}} 
    & \multicolumn{3}{c}{\textbf{0.022 \scriptsize[0.015, 0.060]}}\\
    
    \midrule
    
    Concept &  $\checkmark$ &  $\times$  &  LoRA & 
    0.886{\scriptsize$\pm$0.003} & 0.004{\scriptsize$\pm$0.002} & 0.188{\scriptsize$\pm$0.026} 
    & 0.949{\scriptsize$\pm$0.006} &  0.411{\scriptsize$\pm$0.072} &  0.448{\scriptsize$\pm$0.049} \\
    
    Concept$+$CCSS (Ours) &  $\checkmark$ &  $\checkmark$ &  LoRA & 
    0.890{\scriptsize$\pm$0.002} & 0.005{\scriptsize$\pm$0.001} & 0.166{\scriptsize$\pm$0.011}
    &  0.955{\scriptsize$\pm$0.001} & 0.433{\scriptsize$\pm$0.010} & 0.430{\scriptsize$\pm$0.014} \\
    
    \rowcolor{purple!10}
    \multicolumn{4}{l|}{
    \textbf{$\Delta_{\mathrm{Tail}}$
    (Concept+CCSS $-$ Concept), LoRA [95\% CI]}
    }
    & \multicolumn{3}{c|}{\textbf{0.001{ \scriptsize[0.000, 0.002]} }} 
    & \multicolumn{3}{c}{\textbf{0.023 \scriptsize[0.009, 0.060]}} \\
    
    \bottomrule
    \end{tabular}
    }
\end{table}

\subsubsection{Does \CCSS{} itself matter?}
\label{subsec:ccss-ladder}

Within each adaptation setting, Concept and Concept$+$CCSS use the same backbone, semantic supervision, classifier, training schedule, and seeds. Thus, their difference isolates coverage-aware semantic sharing. As shown in Table~\ref{tab:ccss-ladder}, adding \CCSS{} consistently improves Tail representation quality and reduces the Head--Tail gap under both frozen and LoRA-adapted DINOv3 encoders. Tail Macro-F1 improves by $+0.002$ and $+0.001$ on CUB, and by $+0.022$ and $+0.023$ on CelebA, under frozen and adapted encoders, respectively. Task performance is preserved or slightly improved.

These controlled comparisons separate the effect of \CCSS{} from concept supervision and backbone adaptation. Concept supervision alone does not remove the coverage-dependent representation gap; explicitly adapting semantic sharing to class and concept support and uncertainty provides additional gains for low coverage pairs. This supports the proposed principle: low coverage or uncertain \ccpairs{} benefit from shared semantic structure, while high coverage pairs retain more pair-specific structure.


\subsubsection{Comparison with Existing Methods}
\label{subsec:sota-comparison}


\begin{table}[t]
    \centering
    \small
    \setlength{\tabcolsep}{4pt}
    \renewcommand{\arraystretch}{1.03}
    \caption{
    \textbf{Task-level comparison with existing methods.}
    We report the standard task metric for each dataset: CelebA BAcc, CUB Top-1 accuracy, and AUROC for Derm7pt and MILK10k. Higher is better.
    }
        
    \label{tab:sota-comparison}

    {
    \begin{tabular}{lcccc}
    \toprule
    Method & CelebA & CUB & Derm7pt & MILK10k \\
    \midrule
    JTT  
    & 0.887{\scriptsize$\pm$0.004} 
    & 0.890{\scriptsize$\pm$0.004}  
    & 0.852{\scriptsize$\pm$0.008} 
    & 0.879{\scriptsize$\pm$0.019} \\

    DFR  
    & 0.933{\scriptsize$\pm$0.001} 
    & 0.837{\scriptsize$\pm$0.002}
    & \textbf{0.866{\scriptsize$\pm$0.011}} 
    & 0.859{\scriptsize$\pm$0.017} \\

    MORE 
    & 0.925{\scriptsize$\pm$0.005} 
    & \textbf{0.892{\scriptsize$\pm$0.001}}   
    & 0.854{\scriptsize$\pm$0.009} 
    & 0.879{\scriptsize$\pm$0.013} \\

    TS-MOF 
    & 0.907{\scriptsize$\pm$0.013} 
    & 0.885{\scriptsize$\pm$0.001} 
    & 0.858{\scriptsize$\pm$0.003} 
    & 0.878{\scriptsize$\pm$0.014}  \\

    IVQ-CBM 
    & 0.932{\scriptsize$\pm$0.002} 
    & 0.881{\scriptsize$\pm$0.008}   
    & 0.870{\scriptsize$\pm$0.005} 
    & 0.861{\scriptsize$\pm$0.015}  \\

    \SemCovNet{} (Ours) 
    & \textbf{0.955{\scriptsize$\pm$0.001}} 
    & 0.891{\scriptsize$\pm$0.002} 
    & 0.851{\scriptsize$\pm$0.018}
    & \textbf{0.887{\scriptsize$\pm$0.009}} \\
    \bottomrule
    \end{tabular}
    }
\end{table}

We compare \SemCovNet{} with SOTA methods using the standard task metric for each dataset: Balanced Accuracy for CelebA, Top-1 accuracy for CUB, and AUROC for Derm7pt and MILK10k because each dataset defines different prediction tasks and evaluation protocols. 
Table~\ref{tab:sota-comparison} shows that coverage-aware learning remains competitive at the task level while targeting a different failure mode: under-represented \cc{} semantics. \SemCovNet{} achieves the strongest performance on CelebA and is competitive on CUB, indicating that addressing SCI 
need not trade off aggregate recognition performance. 
We provide a detailed analysis of spurious-correlation robustness on Waterbirds in Appendix~\ref{app:waterbirds-robustness}.

In the medical domain, Derm7pt provides expert-defined clinical concepts, whereas MILK10k provides continuous soft clinical evidence.
\SemCovNet{} remains competitive on Derm7pt and achieves the strongest task performance on MILK10k.
Derm7pt is particularly challenging, with only 413 training samples and
expert-defined clinical concepts whose visual manifestations can be subtle and variable. Its small size and clinical heterogeneity make pair-specific specialization harder to learn and can increase variance across runs.
In this high-stakes medical setting, however, aggregate task performance alone can hide weak representations of rare clinical concepts; a model can perform well overall while still learning weak representations for low coverage \ccpairs. 
This motivates our separate Tail and Head analyses, particularly in clinical settings where rare semantic evidence can be important despite limited effect on aggregate AUROC (see details in Appendix~\ref{app:medical-semantic-supervision}).

\subsection{Controlled Coverage Interventions}
\label{subsec:controlled-intervention}

\begin{wrapfigure}{r}[0pt]{0.48\linewidth} 
    \centering
    \includegraphics[width=\linewidth]{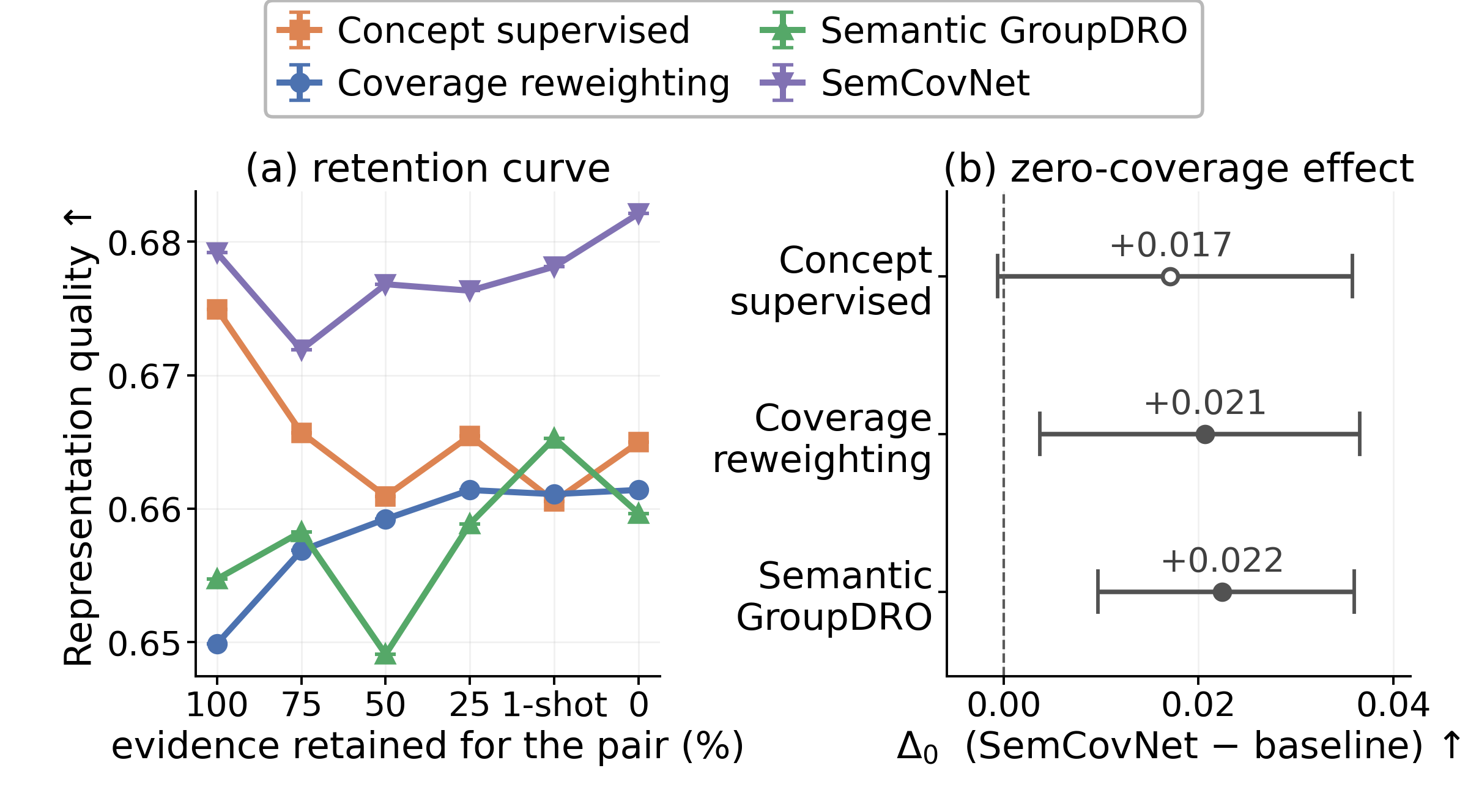}
    \caption{
    \textbf{Controlled \cc{} coverage intervention on CUB.}
    \textbf{(a)} Frozen-probe representation quality as training evidence for target \ccpairs{} is reduced from $100\%$ to $0\%$; ``one'' retains one positive example.
    \textbf{(b)} Paired zero-coverage effect $\Delta_0$ positive values indicate higher representation quality for \SemCovNet{}. Error bars show paired 95\% CIs over target pairs. At zero coverage, \SemCovNet{} retains higher representation quality than Coverage reweight and Semantic GroupDRO.
    }
    \label{fig:fig6-interventions}
\end{wrapfigure}

We next test whether reducing \cc{} coverage degrades semantic representation quality, and whether coverage-aware learning mitigates this effect.
We select CUB because it exhibits the strongest coverage sparsity (\Figref{fig:fig4a-sci-distribution}(b)) and a clear coverage-dependent difference in representation dispersion (\Figref{fig:fig4b-coverage-representation}(c)).
We intervene on selected CUB \ccpairs{} by retaining $100\%$, $75\%$, $50\%$, $25\%$, one example, or $0\%$ of their pair-specific training evidence.
We intervene only on the target pair: its class and concept remain observed through other training pairs, and the test set is unchanged. We measure representation quality using the same frozen-probe protocol as in \Secref{subsec:rep-quality}, with all comparisons paired over the same target pairs. Appendix~\ref{app:controlled-intervention} provides full target-pair selection, intervention details, and results.

\Figref{fig:fig6-interventions}(a) shows small change in \SemCovNet{} representation quality as pair-specific evidence is removed, from $0.679$ at full coverage to $0.682$ at zero coverage. The zero-coverage setting provides the strongest test: the target \ccpair{} is absent from training, while its constituent class and concept remain observed elsewhere. At zero coverage, \SemCovNet{} improves $q_{\mathrm{repr}}$ over Coverage Reweight by $+0.021$ (95\% CI $[+0.004,+0.037]$) and over Semantic GroupDRO by $+0.022$ ($[+0.010,+0.036]$). The point estimate over standard concept supervision is also positive ($+0.017$), although its 95\% CI includes zero ($[-0.001,+0.036]$).

These interventions strengthen the observational result in \Secref{subsec:rep-quality}: reducing pair-specific coverage degrades the evidence available to learn that relation, while \SemCovNet{} remains more robust when direct pair evidence is absent. Reweighting or group-robust objectives can only act on observed training evidence, whereas \CCSS{} can rely more strongly on semantic structure shared across the class and concept when pair-specific support is limited. The zero-coverage comparison therefore supports the proposed coverage-aware learning principle.

\subsection{Coverage-aware Learning Across Clinical Semantic Supervision}
\label{sec:clinical-semantic-supervision}

\begin{wrapfigure}{r}[0pt]{0.48\linewidth} 
    \centering
    \includegraphics[width=0.95\linewidth]{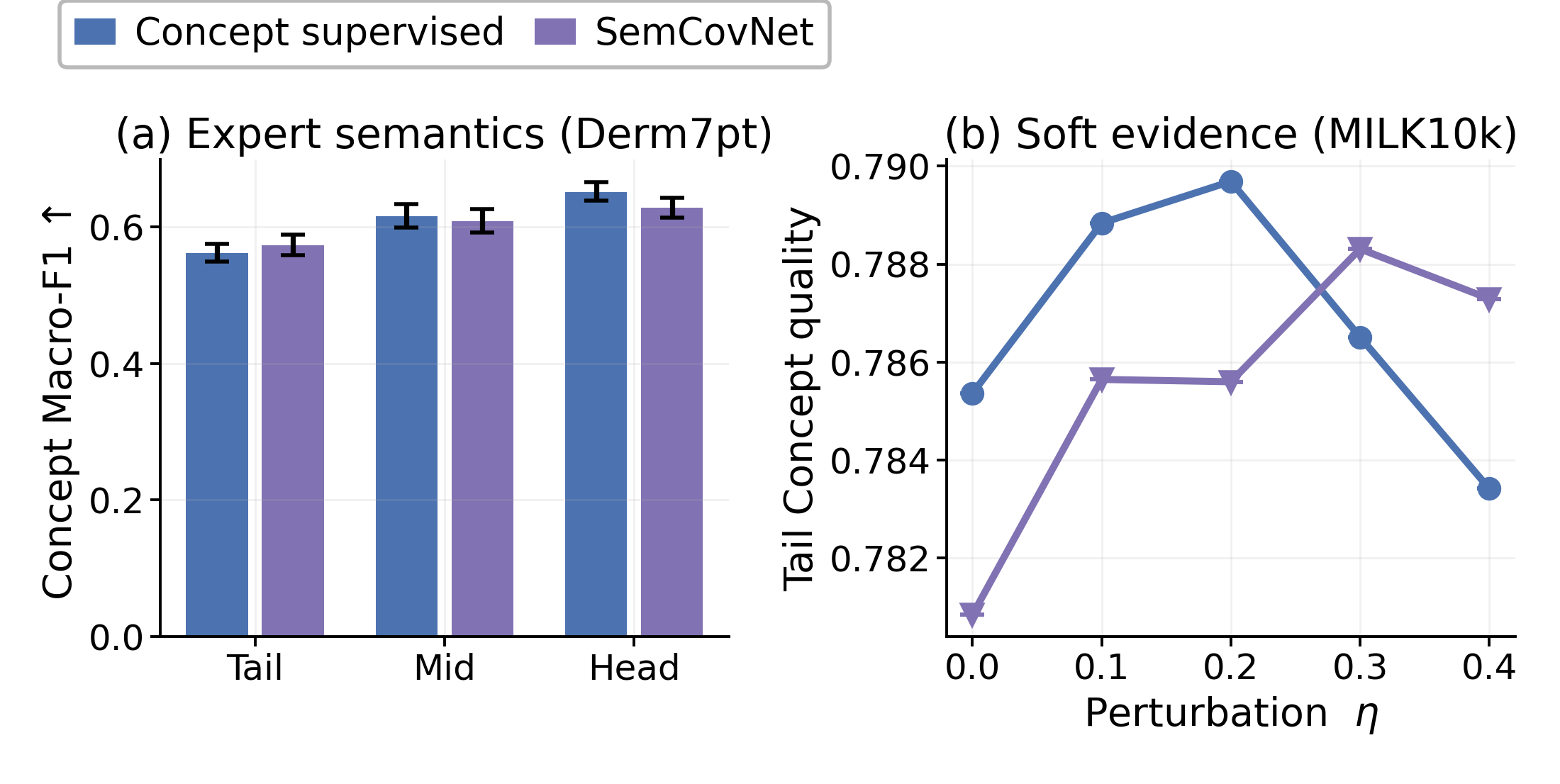}
    \caption{
    \textbf{Coverage-aware learning under clinical semantic supervision.}
    \textbf{(a)} On Derm7pt expert-defined clinical concepts, \SemCovNet{} improves Tail representation quality and reduces the Head--Tail gap over Concept supervision.
    \textbf{(b)} On MILK10k continuous soft semantic supervision, Tail representation quality remains stable as semantic evidence becomes less informative. Error bars show 95\% CIs.
    }
    \label{fig:fig7-medical-generalization}
\end{wrapfigure}

We next evaluate coverage-aware learning under clinical semantic supervision. 
Derm7pt contains only 413 training samples with expert-defined clinical concepts under the 7-point melanoma gold-standard setting. Figure~\ref{fig:fig7-medical-generalization}(a) shows that representation quality remains coverage-dependent. 
\SemCovNet{} improves Tail class-conditioned concept Macro-F1 from $0.562$ to $0.573$ and reduces the Head--Tail gap from $0.090$ to $0.055$, with the main gain concentrated on low coverage pairs.

We further evaluate MILK10k, where semantic supervision is provided as continuous soft evidence scores. We perturb the continuous training evidence to make the semantic target less informative. As \Figref{fig:fig7-medical-generalization}(b) shows, Tail representation quality and task AUROC remain stable over $\eta\in[0,0.4]$. The concept supervised reference is similarly stable, so this experiment tests the applicability of coverage-aware learning to continuous and uncertain semantic evidence rather than uniform robustness. Together, these results show that the same coverage-aware principle applies across expert-defined and soft clinical semantic supervision. Full results and perturbation details are provided in Appendix~\ref{app:medical-semantic-supervision}.

\section{Conclusion}
We study semantic supervision at the class--concept relation level and formalize uneven empirical \ccpair{} coverage as \newterm{Semantic Coverage Imbalance (\SCI{})}.
Across diverse vision datasets, we find that lower \cc{} coverage is associated with weaker semantic representations, motivating coverage-aware sharing of semantic structure for low coverage or uncertain pairs.
Our proposed principle, \newterm{Coverage-aware Semantic Representation Learning}, mitigates this effect, \textit{even} when training evidence for a target \ccpair{} is absent. These findings also show that strong task performance can hide weak representations for low coverage semantic relations.


The broader implication is that semantic supervision should not be evaluated only by which concepts are annotated or how frequent they are globally. What matters is also how semantic evidence is distributed across \ccpairs{}. This perspective is especially relevant in settings such as medical imaging, where insufficient but clinically meaningful relations may receive minimal training evidence. By explicitly considering \ccpair{} coverage, SCI{} provides a new lens for analyzing semantic supervision, while coverage-aware learning shows a principled way to reduce the resulting representation gaps.
This work shows that SCI and coverage-aware semantic representation learning provide a valuable perspective for learning more reliable semantic representations under uneven coverage.

{
    \bibliography{iclr2027_conference,refs-hidden-bias,refs-ltd-reliable,refs-ltd}
    \bibliographystyle{iclr2027_conference}
}

\newpage
\appendix
\section{Appendix}

\addtocontents{toc}{\protect\setcounter{tocdepth}{3}} 
\begingroup
\renewcommand{\contentsname}{} 
\tableofcontents
\endgroup

\newpage

\section{Extended Related Work}
\label{app:related_work}


\textbf{Concept-based visual representation learning.}
Concept Bottleneck Models (CBMs) make predictions through human-interpretable concepts~\citep{koh2020cbm}. Recent work improves concept discovery, grounding, and scalability using vision--language models and richer visual representations~\citep{oikarinen2023labelfree,yuksekgonul2023posthoc,srivastava2024vlgcbm,tan2024opencbm,liu2025hybrid,prasse2025dcbm}.
Fine-grained methods model local or class-specific concept structure: DOT-CBM aligns image patches with concepts~\citep{xie2025dotcbm}, ALBM categorizes concepts in class-specific attribute spaces~\citep{zhang2025albm}, SALF-CBM spatially grounds concept evidence~\citep{benou2025show}, and IVQ-CBM studies many-to-many visual--concept alignment and representation collapse~\citep{gao2026escaping}. 
\citet{torralba2011unbiased} show that visual semantics are far from uniform: modern datasets remain distinguishable through transferable semantic signals~\citep{zeng2024understanding,liu2025a}, and ConceptScope shows uneven class-wise distributions of target, context, and bias concepts~\citep{choi2026conceptscope}. These works treat concepts as useful semantic signals for representation and dataset analysis, but do not show how the \textit{amount of training evidence} supporting each \cc{} relation affects the learned representation.

\textbf{Learning under class and attribute imbalance.}
Long-tailed learning addresses unequal class frequencies through resampling, reweighting, margin or logit adjustment, decoupled training, contrastive learning, and model
adaptation~\citep{cui2019class,cao2019learning,Kang2020Decoupling,menon2021longtail,park2022majority,luo2026long}.
Generalized long-tail learning further shows that attributes can be imbalanced \textit{within} a
class~\citep{tang2022invariant,zhou2025class}, and \citet{zhong2023understanding} show that imbalance can distort feature geometry and reduce minority-class discrimination. 
Group-robust methods instead target poorly performing subpopulations through worst-group optimization, error-based reweighting, or classifier
retraining~\citep{sagawa2020groupdro,nam2020learning,liu2021jtt,kirichenko2023last,jeong2025multiexpert}.
These approaches work primarily over classes, attributes, samples, or predefined groups, rather than the empirical coverage of individual \ccpairs{}. In contrast, \SemCovNet{} uses the effective training support of each \ccpair{}, together with semantic uncertainty, to determine how strongly its representation shares structure across classes and concepts.

\textbf{From semantic imbalance to coverage-aware learning.}
Concept-level failures also appear in learned models: generative models can systematically suppress or exaggerate concepts relative to natural
data~\citep{bohacek2026uncovering}, and concept-based classifiers can suffer from incomplete supervision, noisy concepts, or collapsed representations~\citep{penaloza2025cpo,park2025an,hu2025semi,gao2026escaping}.
We study a more inherent source of failure: uneven empirical coverage across \ccpairs{} in the training data. 
Under \SCI{}, class and concept marginals may appear adequate while particular \ccpairs{} have limited effective support, yielding noisier semantic representations.
This motivates \textit{coverage-aware semantic representation learning}: low coverage or uncertain pairs should share more semantic structure, while high coverage pairs should retain more pair-specific structure. \SemCovNet{} instantiates this principle through \CCSS{} and combines the resulting concept representations with a residual visual pathway.

\textbf{Limits of concept supervision.}
Beyond CBMs, recent work studies whether a concept vocabulary is complete or well-specified.
\citet{luyten2024conceptsets} formalize what makes a concept set good, its expressiveness and its inductive bias for the downstream model, and show that a well-chosen set improves sample efficiency and out-of-distribution robustness, motivating a method for selecting informative, non-redundant concepts. 
\citet{zhang2023decoupling} show that an insufficient concept set forces a bottleneck to blend explicit and implicit information, distorting both the learned concepts and the final label; their Decoupling CBM separates this mixture into explicit and implicit branches to recover accuracy without abandoning interpretability.
SCI studies a different question, also distinct from prior dataset-bias work~\citep{torralba2011unbiased,zeng2024understanding,liu2025a,bohacek2026uncovering,choi2026conceptscope}: even with a complete, well-specified vocabulary, a \ccpair{} can still receive abundant evidence in one class and almost none in another, and we study whether this unequal distribution affects the representation learned for that pair.

\textbf{Subgroup robustness and semantic coverage.}
Group-robust methods correct for imbalance at a specific unit of analysis, and each implicitly assumes that unit has enough evidence to be estimated reliably. GroupDRO~\citep{sagawa2020groupdro} minimizes a minimax objective over predefined groups, so it needs the group label in advance and enough examples per group to estimate each group's loss reliably.
 JTT~\citep{liu2021jtt} avoids predefined groups by first training a standard model and upweighting the examples it misclassifies, using early mistakes as a proxy for minority-group membership; this proxy is only informative if enough misclassified examples exist to upweight. 
DFR~\citep{kirichenko2023last} instead freezes the features from standard training and retrains only the last layer on a small group-balanced sample, which shows the features already carry the signal but still requires a held-out set large enough to balance every group. 
In each case, imbalance is corrected at the level of a group, a misclassified example, or a balanced held-out set but not at the level of \cc{} relation. A \ccpair{} with very little support can therefore fall inside a group, or a balanced sample, that looks adequate on average, while the pair itself remains of minimum support and hidden from these objectives.

\section{Experimental Setup}
\label{sec:exp-setup}






\subsection{Datasets and Semantic Evidence}
\label{app:datasets}

We evaluate across five visual datasets spanning natural images, fine-grained recognition, spurious correlations, and medical imaging. The datasets provide different forms of semantic evidence, allowing us to test whether \SCI{} extends beyond a single annotation type. Table~\ref{tab:sci-dataset-stat} reports dataset-level coverage statistics. In this study, we compute coverage $C_{y,k}$ only from the training split.

\textbf{CelebA}
\citep{liu2015faceattributes} provides binary facial attributes and is used as a general visual-recognition benchmark. CelebA contains 162,770 training and 19,962 test images of celebrity faces annotated with 40 binary attributes. We treat \texttt{blond\_hair} as the binary classification target and use the remaining 39 attributes as concept sources. We use the 39 available attributes as atomic semantic concepts. Therefore, $\eva_{i,k}\in{0,1}$ indicates whether concept $k$ is present in image $\vx_i$. This setting evaluates SCI under dense binary attribute supervision.

\textbf{CUB}
\citep{wah2011caltech} contains 200 bird species with 312 annotated visual attributes. We treat the 312 attributes as atomic concepts and use the 28 attribute families only as semantic metadata. All coverage computation, representation learning, and pair-level analysis operate on atomic \ccpairs{}. CUB is particularly useful for studying sparse and zero-coverage relations because many species--attribute combinations obtain small or no direct training evidence.

\textbf{WaterBirds}
\citep{sagawa2020groupdro} is included as a complementary spurious-correlation benchmark. We use its standard class and group annotations to evaluate robustness under background correlation. Its role differs from the primary SCI characterization: it tests whether coverage-aware semantic learning remains competitive in a setting commonly studied through group robustness.

\textbf{Derm7pt}
\citep{Kawahara2018-7pt} associates skin-lesion diagnoses with gold-standard expert-defined clinical descriptors for the 7-point melanoma checklist. We expand categorical clinical criteria into atomic semantic concepts, yielding 28 concepts organized into seven clinical descriptors. The descriptors are used only for organization; \ccpairs{} remain the primary unit of coverage and representation analysis. Derm7pt therefore evaluates SCI with expert-defined clinical semantic supervision.

\textbf{MILK10k}
\citep{philipp2025milk10k} provides continuous semantic evidence rather than binary concept labels. MILK10k officially provides seven MONET-driven descriptor soft scores and uses them as soft supervision during training.
This setting tests whether coverage-aware learning extends to uncertain and continuous semantic supervision.

\subsection{Baseline Methods}
\label{app:baselines}

We compare \SemCovNet{} with established and recent methods from foundation-model representations, imbalance-aware learning, group robustness, and concept-based representation learning. We use each baseline to test a different explanation for weakly under-covered semantics.

\textbf{Representation and last-layer baselines.}
DINOv3--ERM uses DINOv3~\citep{simeoni2025dinov3} features with a linear prediction head and provides the main no-coverage-awareness control. DFR~\citep{kirichenko2023last} retrains the final classifier on fixed features and tests whether a strong representation already contains the information needed for robust prediction. These baselines separate representation quality from specialized semantic learning.

\textbf{Reweighting and robust learning.}
JTT~\citep{liu2021jtt} emphasizes examples that are misclassified by an initial model and tests whether error-driven reweighting is sufficient. GroupDRO~\citep{sagawa2020groupdro} directly optimizes worst-group performance. We additionally consider recent representation- and distribution-level robustness methods, including SCER~\citep{park2026spurious}, which regularizes embeddings against spurious correlations, and Hierarchical DRO~\citep{jo2026mitigating}, which models both inter-group and intra-group distributional uncertainty. 

\textbf{Long-tail learning.}
MORE~\citep{luo2026long} addresses long-tailed recognition by reallocating model capacity toward underrepresented classes, while TS-MOF~\citep{zhao2025tsmof} performs multi-objective classifier adaptation over frozen representations. These comparisons test whether methods designed for frequency imbalance also address the class-conditioned semantic coverage problem studied here.

\textbf{Concept-based learning.}
We compare with recent concept-learning approaches where applicable. IVQ-CBM~\citep{gao2026escaping} learns high-rank visual concept representations through implicit vector quantization. 
These methods test whether stronger concept representations, concept robustness, or structured concept learning can resolve SCI.

\textbf{Controlled baselines and ablations.}
\textit{Concept Supervision} uses the same semantic supervision without CCSS. \textit{Coverage Reweighting} adjusts the contribution of low coverage semantic relations but does not borrow shared semantic structure. \textit{Semantic GroupDRO} applies group-robust optimization~\citep{sagawa2020groupdro} to semantic groups or pairs, testing whether worst-group optimization alone is sufficient. Finally, we compare \textit{Frozen} and \textit{LoRA} variants while keeping the DINOv3 encoder fixed; the latter applies parameter-efficient adaptation to the final three transformer blocks. Within each controlled comparison, we match the backbone, supervision, classifier, training schedule, and seeds; only the tested mechanism changes.

\subsection{Evaluation Protocol}
\label{app:evaluation-protocol}

\textbf{Task performance.}
We use the standard task metric for each benchmark. We report balanced accuracy (BAcc) on CelebA, Top-1 accuracy on CUB, average accuracy on WaterBirds, and AUROC on Derm7pt and MILK10k. BAcc accounts for class imbalance in CelebA, Top-1 accuracy is the standard metric for CUB's multi-class recognition task, and AUROC provides a threshold-independent measure for the medical classification tasks. Using the task-specific metric for each dataset also allows comparison with prior work.

\textbf{Tail/Middle/Head Coverage pairs.}
For semantic analysis, we rank \ccpairs{} by their training coverage $C_{y,k}$  and use the 25th and 75th percentiles to define
$$
    \text{Tail}:Q_{0\text{--}25\%},
    \qquad
    \text{Middle}:Q_{25\text{--}75\%},
    \qquad
    \text{Head}:Q_{75\text{--}100\%}.
$$
We compute the Tail/Middle/Head pairs once from the training split and fix them for evaluation. We use them only for reporting and visualization; our learning method does not use them as thresholds. We use the official test set only to evaluate the representation learned for each training-defined \ccpair{}.

\textbf{Semantic representation quality.}
Class--concept pair-level semantic analyses use the learned concept representation $\vz^{\mathrm{sem}}$. We report class-conditioned concept Macro-F1 for Tail, Middle, and Head pair representation quality and the Head--Tail gap $\Delta_{\mathrm{H-T}}=\mathrm{Head}-\mathrm{Tail}.$
Frozen linear probes measure how well each concept is represented without updating the encoder. Unless stated otherwise, we aggregate pair-level probe results over \ccpairs{} within each training coverage pair. We use visual concept tokens $z_{i,k}$ for semantic representation analyses.

\textbf{Task and diagnostic representations.}
We compute task performance from the model's saved predictions or logits rather than from concept probes. When a task-level linear probe is reported, we use the joint semantic--visual representation $\vh_i$. 
DINOv3-ERM baseline is evaluated on its global feature for task performance and its class token \texttt{CLS} representation. This separation keeps task performance distinct from how well individual semantic representations perform.

\textbf{95\% CIs.}
We report mean performance across training seeds and use 95\% bootstrap confidence intervals over \ccpairs{} for pair-level analyses. For controlled pair interventions, the \ccpair{} is the experimental unit: we first aggregate seed results within each pair, then compute confidence intervals across pairs.

\subsection{Implementation Details}
\label{app:implementation-details}

We use DINOv3 ViT-B/16 as the primary visual encoder. In the frozen setting, we keep the pretrained encoder fixed, while in the parameter-efficient setting, we apply LoRA to the final three transformer blocks. The semantic branch produces concept-specific representations $z_{i,k}$, and the model retains a residual visual representation so that prediction is not restricted to the predefined concept vocabulary. 

We instantiate $\ell_{\mathrm{sem}}$ (\Eqref{eq:task_concept_losses}) based on the specific dataset's concept annotations: binary cross-entropy for binary concepts, categorical supervision for grouped categorical attributes where applicable, and a soft-target loss for continuous semantic evidence.

For controlled comparisons, we use identical data splits, intervention manifests, initialization seeds, and evaluation protocol across methods. Concept and Concept$+$CCSS variants also share the same backbone, semantic supervision, classifier, and optimization schedule, so their difference isolates CCSS.

All Tail/Middle/Head coverage pairs are computed from the training split only. Dataset validation and test splits are never used to estimate coverage or select intervention pairs. We select hyperparameters using validation data and use the test split only for final evaluation. Table~\ref{tab:implementation-details} presents the optimizer, learning rates, batch sizes, number of epochs, LoRA configuration, loss weights, input resolution, and dataset-specific preprocessing settings. 

\begin{table}[!t]
    \centering
    \setlength{\tabcolsep}{5pt}
    \renewcommand{\arraystretch}{1.05}
    \caption{
        \textbf{Implementation details.}
        Methods use the same backbone, data splits, optimization settings, and random seeds.
    }
    \label{tab:implementation-details}
    \begin{tabular}{l l}
        \toprule
        \textbf{Setting} & \textbf{Configuration} \\
        \midrule
        Visual encoder      & DINOv3 ViT-B/16 \\
        Input resolution    & \texttt{224$\times$224} \\
        Encoder adaptation  & Frozen / LoRA on final 3 transformer blocks \\
        LoRA rank $r$       & \texttt{8} \\
        LoRA scaling $\alpha$ & \texttt{16} \\
        LoRA dropout        & \texttt{0.05} \\
        Optimizer           & \texttt{AdamW} \\
        Learning rate       & \texttt{3e-4} \\
        Weight decay        & \texttt{0.05} \\
        Batch size          & \texttt{64} \\
        Training epochs     & \texttt{20} \\
        Concept-loss weight & \texttt{1.0} \\
        CCSS-loss weight    & \texttt{0.25} \\
        Coverage source     & Training split only \\
        Semantic unit       & class--concept pair $(y,k)$ \\
        \bottomrule
    \end{tabular}
\end{table}

\section{Additional Formalization and Method Details}
\label{app:formalization-method}

\subsection{Mathematical Characterization of Semantic Coverage Imbalance}
\label{app:sci-definition}

This section formally characterizes of \newterm{Semantic Coverage Imbalance (\SCI{})} and clarifies its relationship to effective support, class and concept marginals, as well as the coverage pairs used for analysis.

\paragraph{Class--concept coverage.}
Following Definition~\ref{def:cccov}, let
\(
    \train=\{(\vx_i,y_i,\va_i)\}_{i=1}^{n}
\)
denote the training set, where $a_{i,k}\in[0,1]$ is the semantic evidence
for concept $k$ in image $\vx_i$.
For class $y$, let
\[
    \train_y:=\{\vx_i:y_i=y\},
    \qquad
    n_y:=|\train_y|.
\]
The empirical coverage of \ccpair{} $(y,k)$ is
\begin{equation}
    C_{y,k}
    =
    \frac{1}{n_y}
    \sum_{i:y_i=y} a_{i,k},
    \qquad
    C_{y,k}\in[0,1].
    \label{eq:app-coverage}
\end{equation}
Using the effective support from \Eqref{eq:n_effective_support},
\[
    n^{\mathrm{eff}}_{y,k}
    =
    \sum_{i:y_i=y} a_{i,k},
\]
we equivalently have
\begin{equation}
    C_{y,k}
    =
    \frac{n^{\mathrm{eff}}_{y,k}}{n_y}.
    \label{eq:coverage-support-relation}
\end{equation}
Thus, coverage measures the observed support for relation $(y,k)$ relative to
the size of class $y$, while effective support measures its total observed
evidence.

\paragraph{Coverage distribution.}
Let
\begin{equation}
    \gP
    :=
    \{(y,k)\in\sY\times\sK:n_y>0\}
    \label{eq:valid-pairs}
\end{equation}
denote the valid \ccpairs{} in the training set, including pairs with zero
observed semantic evidence.
Their empirical coverage distribution is
\begin{equation}
    \mathcal{C}_{\train}
    :=
    \{C_{y,k}:(y,k)\in\gP\}.
    \label{eq:coverage-distribution}
\end{equation}
We deliberately do not define \SCI{} using a fixed coverage threshold.
Instead, its severity and structure are characterized by the empirical distribution $\mathcal{C}_{\train}$ itself.

\paragraph{Why class and concept marginals can hide \SCI{}.}
Let
\(
    p_y := n_y/n
\)
denote the empirical class frequency and
\(
    \bar{a}_k := \frac{1}{n}\sum_{i=1}^{n} a_{i,k}
\)
the global prevalence of concept $k$.
From \Eqref{eq:app-coverage},
\begin{equation}
    \bar{a}_k
    =
    \sum_{y\in\sY}
    p_y\,C_{y,k}.
    \label{eq:concept-marginal-decomposition}
\end{equation}
Thus, global concept prevalence is a class-frequency-weighted average of
class-conditioned coverage.
Different distributions of $\{C_{y,k}\}_{y\in\sY}$ can therefore produce the
same marginal prevalence, allowing low coverage \cc{} relations to remain
hidden by aggregate class and concept statistics.
\paragraph{Ranked coverage and the low coverage tail.}
Let
\begin{equation}
    C_{(1)}
    \ge
    C_{(2)}
    \ge
    \cdots
    \ge
    C_{(N)},
    \qquad
    N:=|\gP|,
    \label{eq:ranked-coverage}
\end{equation}
denote the coverage values sorted in decreasing order.
The resulting curve summarizes the distribution of empirical \ccpair{}
coverage and its low coverage tail.

For a descriptive level $\delta\in[0,1]$, we define the low coverage tail mass
as
\begin{equation}
    p_{<\delta}
    =
    \frac{1}{|\gP|}
    \sum_{(y,k)\in\gP}
    \mathbf{1}\!\left[C_{y,k}<\delta\right],
    \label{eq:sci-tail-mass}
\end{equation}
and the zero-coverage fraction as
\begin{equation}
    p_{0}
    =
    \frac{1}{|\gP|}
    \sum_{(y,k)\in\gP}
    \mathbf{1}\!\left[C_{y,k}=0\right].
    \label{eq:sci-zero-mass}
\end{equation}

\begin{remark}[Descriptive coverage groups]
\label{rem:coverage-groups}
Descriptive levels $\delta$ summarize low coverage tail mass, while the
dataset-specific 25th and 75th coverage quantiles define the Tail
($Q_{0\text{--}25\%}$), Middle ($Q_{25\text{--}75\%}$), and Head
($Q_{75\text{--}100\%}$) groups used in the evaluation protocol.
These statistics characterize the prevalence and severity of low coverage
relations without defining \SCI{} through a universal threshold.
The coverage groups are used only for evaluation and visualization and are
not used by \SemCovNet{} or \CCSS{} during training.
\end{remark}
    \paragraph{Distinguishing \SCI{} from class imbalance.}
Class imbalance concerns variation in class counts $\{n_y\}_{y\in\sY}$,
whereas \SCI{} concerns variation in \ccpair{} coverage
$\{C_{y,k}\}_{(y,k)\in\gP}$.
The two may coexist, but balancing class counts does not necessarily equalize
\ccpair{} coverage.
We examine this distinction empirically through class-balanced resampling and prevalence-preserving permutation analyses in Appendix~\ref{app:sci-characterization}.

\paragraph{Soft semantic evidence.}
For soft semantic evidence, $\eva_{i,k}\in[0,1]$ denotes the confidence score
associated with concept $k$ in image $\vx_i$.
The definitions of coverage and effective support remain unchanged:
\[
    C_{y,k}
    =
    \frac{1}{n_y}\sum_{i:y_i=y}\eva_{i,k},
    \qquad
    n^{\mathrm{eff}}_{y,k}
    =
    \sum_{i:y_i=y}\eva_{i,k}.
\]
Thus, each observation contributes according to its evidence score rather than
as a hard binary count.

\subsection{Coverage-Calibrated Semantic Sharing: Derivation and Diagnostics}
\label{app:ccss}

\textbf{Hierarchical model and posterior derivation.}
CCSS follows from a hierarchical Gaussian model for the latent coverage-calibrated representation $\hat{\vt}_{y,k}$ of each \ccpair{}:
\begin{align}
    \hat{\vt}_{y,k} &\sim \mathcal{N}(\vm_{y,k},\, \tau^2\mI),
    \qquad
    \bar{\vz}_{y,k}\mid \hat{\vt}_{y,k}
    \sim
    \mathcal{N}\!\left(\hat{\vt}_{y,k},\, \frac{\sigma_k^2}{n^{\mathrm{eff}}_{y,k}}\mI\right),
    \label{eq:hier_obs}
\end{align}
where $\vm_{y,k}$ is the shared prior mean (\Eqref{eq:shared_semantic_prior}); $\tau^2$ is how far a true pair representation can drift from that prior; $\sigma_k^2/n^{\mathrm{eff}}_{y,k}$ is the sampling variance of $\bar{\vz}_{y,k}$, built from $n^{\mathrm{eff}}_{y,k}$ effective observations. Under this Gaussian prior and likelihood, the posterior mean of $\hat{\vt}_{y,k}$ given $\bar{\vz}_{y,k}$ is exactly \Eqref{eq:ccss_target}, with $\alpha_{y,k}$ in \Eqref{eq:ccss_weight} the corresponding shrinkage weight. \Eqref{eq:ccss_target} is therefore the Bayes-optimal estimate of $\hat{\vt}_{y,k}$ under this model, for any $\tau^2,\sigma_k^2>0$, not a heuristic interpolation.

\paragraph{Estimating between-pair variance $\tau^2$.}
The residual
$\bar{\vz}_{y,k}-\vm_{y,k}$ contains both genuine between-pair variation and
sampling noise due to finite effective support.
Under \Eqref{eq:hier_obs},
\[
\E\!\left[
    \|\bar{\vz}_{y,k}-\vm_{y,k}\|_2^2
\right]
=
D\tau^2
+
\frac{D\sigma_k^2}{n^{\mathrm{eff}}_{y,k}} .
\]
We therefore subtract the sampling-noise term and average over valid pairs
$\sV$ to estimate $\tau^2$, as in \Eqref{eq:tau2_estimator}.
Because an individual corrected residual can be negative due to sampling
noise, $[\cdot]_+$ enforces the non-negativity of the final variance estimate.

\paragraph{Why pool $\sigma_k^2$ at the concept level.}
Estimating a separate $\sigma_{y,k}^2$ for every \ccpair{} is unreliable
precisely when $n^{\mathrm{eff}}_{y,k}$ is small.
Instead, \Eqref{eq:concept_uncertainty} pools evidence across classes to
estimate concept-level uncertainty $\sigma_k^2$.
This provides a more stable uncertainty estimate for low-support pairs, where
\CCSS{} relies more strongly on shared semantic structure.
We evaluate pair-specific uncertainty $\sigma_{y,k}^2$ as an ablation in
Appendix~\ref{app:ablation-alpha}.

\paragraph{Why use a global $\tau^2$.}
Estimating a separate between-pair variance for each concept can be unstable
when that concept has limited coverage across classes.
We therefore use a single global $\tau^2$, estimated across valid \ccpairs{},
to capture the overall scale of pair-specific variation around the shared
semantic prior.
This yields a stable calibration signal even when individual concepts occur
in only a few classes.
\paragraph{Zero-support pairs.}
When $n^{\mathrm{eff}}_{y,k}=0$, \Eqref{eq:ccss_weight} gives
$\alpha_{y,k}=0$, and the calibrated target in
\Eqref{eq:ccss_target} reduces to
\[
    \hat{\vt}_{y,k}=\vm_{y,k}.
\]
Thus, when a \ccpair{} has no observed pair-specific semantic evidence,
\CCSS{} relies entirely on the shared semantic prior.
We use $\eps$ in the relevant denominators and apply $[\cdot]_+$ in
\Eqref{eq:tau2_estimator} for numerical stability and to ensure a
non-negative variance estimate.

\paragraph{Estimation during training.}
The empirical pair representations $\bar{\vz}_{y,k}$ and uncertainty
statistics $\sigma_k^2$ and $\tau^2$ are estimated from the training split as
the concept representations evolve.
These quantities are treated as stop-gradient training statistics when
constructing the calibrated target $\hat{\vt}_{y,k}$ and computing
$\Ls_{\mathrm{CCSS}}$.
Thus, \CCSS{} uses the current representation statistics for calibration
without backpropagating through their estimation.

\paragraph{Normalization order.}
The calibrated target in \Eqref{eq:ccss_target} is formed in the
unnormalized latent space:
\[
    \hat{\vt}_{y,k}
    =
    (1-\alpha_{y,k})\vm_{y,k}
    +
    \alpha_{y,k}\bar{\vz}_{y,k}.
\]
Normalization is applied only when evaluating the cosine objective in
\Eqref{eq:ccss_loss}.
This preserves the intended partial-pooling interpretation: small
$\alpha_{y,k}$ places the target closer to the shared semantic prior, whereas
large $\alpha_{y,k}$ places greater weight on the empirical pair-specific
representation.
We additionally monitor the norms of $\vm_{y,k}$ and
$\bar{\vz}_{y,k}$ to verify that their scales remain comparable.




\subsection{How CCSS Updates the Shared Semantic Structure}
\label{app:ccss_gradient}

The shared semantic structure \Eqref{eq:shared_semantic_prior} is 
\begin{equation*}
    \vm_{y,k}
    =
    \vmu_0
    +
    \tilde{\vmu}_y
    +
    \tilde{\vmu}_k
\end{equation*}
contains learnable global, class-shared, and concept-shared parameters.
These parameters do not receive separate global-, class-, or concept-level supervision. Instead, the instance-level CCSS objective learns them through the \ccpair{} associated with each visual concept token.

For image $i$ with class label $y_i$, the concept token $\vz_{i,k}$ is aligned with the coverage-calibrated target \Eqref{eq:ccss_target} is 
\begin{equation*}
    \hat{\vt}_{y_i,k}
    =
    (1-\alpha_{y_i,k})\vm_{y_i,k}
    +
    \alpha_{y_i,k}\bar{\vz}_{y_i,k}.
\end{equation*}
Thus, the minibatch does not require an additional pair-level label:
the pair is determined directly by the observed class $y_i$ and
concept index $k$, while $a_{i,k}$ determines the contribution of that
concept to the loss. Specifically, \Eqref{eq:ccss_loss} is 
\begin{equation*}
    \Ls_{\mathrm{CCSS}}
    \approx
        \frac{
        \sum_i\sum_k
        a_{i,k}\,
        \ell_{\mathrm{cos}}
        \!\left(
            \vz_{i,k},
            \hat{\vt}_{y_i,k}
        \right)
    }{
        \sum_i\sum_k a_{i,k}+\eps
    }.
\end{equation*}
During CCSS optimization, $\bar{\vz}_{y,k}$ and the statistics used to
compute $\alpha_{y,k}$ are treated as stop-gradient quantities, whereas
$\vm_{y,k}$ remains differentiable. Hence, for a class--concept pair
$(y,k)$,
\begin{equation}
    \vg_{y,k}
    :=
    \frac{\partial\mathcal{L}_{\mathrm{CCSS}}}
         {\partial\vm_{y,k}}
    =
    (1-\alpha_{y,k})
    \frac{\partial\mathcal{L}_{\mathrm{CCSS}}}
         {\partial\hat{\vt}_{y,k}}.
    \label{eq:app_pair_gradient}
\end{equation}
All instances in the minibatch with the same $(y,k)$ contribute to the
same pairwise gradient $\vg_{y,k}$. Because the same parameters are
reused across multiple pairs, these pairwise gradients are then
aggregated according to the sharing structure:
\begin{equation}
    \nabla_{\vmu_0}\mathcal{L}_{\mathrm{CCSS}}
    =
    \sum_{y,k}\vg_{y,k},
    \label{eq:app_global_gradient}
\end{equation}
while $\vmu_y$ receives the aggregate signal from pairs sharing class
$y$, and $\vmu_k$ from pairs sharing concept $k$. With the centered
effects
\(
\tilde{\vmu}_y=\vmu_y-\frac{1}{C}\sum_{c}\vmu_c
\)
and
\(
\tilde{\vmu}_k=\vmu_k-\frac{1}{K}\sum_{j}\vmu_j,
\)
their gradients are
\begin{align}
    \nabla_{\vmu_y}\mathcal{L}_{\mathrm{CCSS}}
    &=
    \sum_k \vg_{y,k}
    -
    \frac{1}{C}
    \sum_{y'}\sum_k \vg_{y',k},
    \\
    \nabla_{\vmu_k}\mathcal{L}_{\mathrm{CCSS}}
    &=
    \sum_y \vg_{y,k}
    -
    \frac{1}{K}
    \sum_{k'}\sum_y \vg_{y,k'}.
\end{align}

Therefore, $\vmu_0$ aggregates information across all observed \ccpairs{}, $\tilde{\vmu}_y$ captures deviations shared across concepts of class $y$, and $\tilde{\vmu}_k$ captures deviations shared across classes for concept $k$. Importantly, this structure is induced by parameter sharing rather than by additional supervision. 
Moreover, the factor $(1-\alpha_{y,k})$ makes sharing adaptive: low coverage (low-support) or uncertain pairs are pulled more strongly toward shared semantic structure, whereas high coverage (well-supported) pairs place greater weight on their empirical pair-specific representation $\bar{\vz}_{y,k}$.

\paragraph{Illustrative minibatch gradient routing.}
No explicit \ccpair{} label is required by the dataloader. For an
instance $(\vx_i,y_i,\va_i)$, the class label $y_i$ and concept index
$k$ select the calibrated target $\hat{\vt}_{y_i,k}$, while
$a_{i,k}$ weights that instance--concept contribution in
$\mathcal{L}_{\mathrm{CCSS}}$.

For example, consider
\[
    \begin{array}{c|c|cc}
        & y_i & a_{i,1} & a_{i,2} \\
    \hline
    i_1 & y_1 & 1 & 1 \\
    i_2 & y_1 & 0 & 1 \\
    i_3 & y_2 & 1 & 0
    \end{array}
\]
which induces the CCSS terms
\[
    \ell(\vz_{1,1},\hat{\vt}_{1,1}),\quad
    \ell(\vz_{1,2},\hat{\vt}_{1,2}),\quad
    \ell(\vz_{2,2},\hat{\vt}_{1,2}),\quad
    \ell(\vz_{3,1},\hat{\vt}_{2,1}).
\]
Hence, the two instances contributing to $(y_1,k_2)$ jointly produce
the pairwise gradient $\vg_{1,2}$, whereas $(y_1,k_1)$ and
$(y_2,k_1)$ produce $\vg_{1,1}$ and $\vg_{2,1}$, respectively.
Autograd then routes these gradients through
\(
\vm_{y,k}=\vmu_0+\tilde{\vmu}_y+\tilde{\vmu}_k
\),
so that, ignoring the centering correction for illustration,
\[
    \vmu_0 \leftarrow \vg_{1,1}+\vg_{1,2}+\vg_{2,1},\qquad
    \vmu_{y_1} \leftarrow \vg_{1,1}+\vg_{1,2},\qquad
    \vmu_{k_1} \leftarrow \vg_{1,1}+\vg_{2,1}.
\]
Therefore, \cc{} pairwise supervision occurs from the instance-level CCSS loss:
$\vmu_0$ aggregates signals across all observed pairs,
$\vmu_y$ across concepts sharing class $y$, and
$\vmu_k$ across classes sharing concept $k$.

\begin{table}[h]
    \centering
    \caption{
    \textbf{Dataset-level characterization of Semantic Coverage
    Imbalance.} 
    Coverage $C_{y,k}$ is computed per \ccpair{} on the training split only. Near-zero and Zero denote the fraction of pairs with $C_{y,k}<0.05$ and $C_{y,k}=0$, respectively. 
    }
    \label{tab:sci-dataset-stat}
    \resizebox{\linewidth}{!}{
    \begin{tabular}{lrrrrccccc}
        \toprule
        Dataset &
        \#train &
        Class &
        Concept &
        Pairs &
        $Q_{25}$ &
        Median &
        $Q_{75}$ &
        Near-zero (\%) &
        Zero (\%) \\
        \midrule
        CelebA & 162770 &
        2 & 39 & 78 &
        0.048 & 0.143 & 0.338 &
        26.9 & 0.0 \\

        CUB &
        5394 & 200 & 312 & 62,400 &
        0.000 & 0.000 & 0.111 &
        67.2 & 53.9 \\

        Derm7pt &
        413 & 2 & 28 & 56 &
        0.031 & 0.117 & 0.455 &
        33.9 & 12.5 \\

        Waterbirds &
        4795 & 2 & 2 & 4 &
        0.050 & 0.500 & 0.950 &
        25.0 & 0.0 \\

        MILK10k &
        3668 & 11 & 7 & 77 &
        0.253 & 0.368 & 0.433 &
        0.0 & 0.0 \\
        \bottomrule
    \end{tabular}}
\end{table}

\begin{table}[h]
    \centering
    \caption{
    \textbf{Statistical controls for \SCI{}.} Significant heterogeneity
    is the share of concepts whose across-class coverage variation
    exceeds a prevalence-preserving permutation null after BH--FDR
    correction ($q<0.05$). $\rho_{\mathrm{bal}}$ is the Spearman correlation between full-data and
class-balanced coverage estimates; ``---'' denotes not applicable.
    }
    \label{tab:sci-controls}
    \begin{tabular}{lcc}
        \toprule
        Dataset & Significant heterogeneity (\%) & $\rho_{\mathrm{bal}}$ \\
        \midrule
        CelebA     & 97.4  & 1.000 \\
        CUB        & 96.5  & 0.996 \\
        Derm7pt    & 64.3  & 0.993 \\
        Waterbirds & 100.0 & 0.800 \\
        MILK10k    & 85.7  & ---   \\
        \bottomrule
    \end{tabular}
\end{table}

\section{Extended Results and Discussion}
\label{app:results}

\subsection{Detailed Characterization of Semantic Coverage Imbalance}
\label{app:sci-characterization}

Tables~\ref{tab:sci-dataset-stat} and~\ref{tab:sci-controls} report the full training-set coverage statistics underlying Fig.~\ref{fig:fig4a-sci-distribution}, including two datasets: Waterbirds and MILK10k. CelebA, CUB, and Derm7pt exhibit substantial variation in class-conditioned concept coverage despite large differences in dataset size, number of classes, and semantic vocabulary. The permutation analysis further shows that the observed across-class heterogeneity is substantially greater than expected from global concept prevalence and finite sampling alone. Class-balanced resampling preserves the original coverage ordering almost perfectly, confirming that the observed semantic structure is not an artifact of target-class counts. Waterbirds and MILK10k also follow the same pattern: substantial across-class coverage heterogeneity (100.0\% and 85.7\% of concepts significant) that class-balanced resampling does not explain ($\rho_{\mathrm{bal}}=0.800$ for Waterbirds; not computed for MILK10k due to its small sample size).

CUB exhibits particularly sparse coverage because its 312 fine-grained attributes form highly structured class--concept compositions; some zero-coverage pairs may therefore represent genuine semantic incompatibility rather than harmful low coverage. We retain all pairs when characterizing the dataset-level coverage structure, but apply predefined support criteria in subsequent representation and performance analyses. Thus, \Secref{sec:results-sci} establishes the existence and structure of SCI without assuming that every naturally rare or absent \ccpair{} constitutes a learning failure.


\subsection{Full Results for Controlled Coverage Interventions}
\label{app:controlled-intervention}

\subsubsection{Setup and Target-Pair Selection}
\label{app:e4-setup}

We conduct the controlled intervention on CUB, where the same semantic concept can occur across multiple bird classes. We first enumerate all \ccpairs{} $(y,k)$ using the training split only. A pair is eligible if it has sufficient positive training support for controlled retention, sufficient test support for stable evaluation, and the corresponding concept $k$ remains observed in other classes. We also require class $y$ to remain represented by other semantic concepts. These constraints confirm that the intervention removes evidence for the \textit{target combination}, rather than removing the class or concept itself. Applying these criteria yields 24 target \ccpairs{}, which are fixed before any intervention or model comparison in this experiment.

For each target pair, we retain $100\%$, $75\%$, $50\%$, $25\%$, one positive example, or $0\%$ of its original training evidence. All non-target samples and the validation/test sets remain unchanged. Therefore, we manipulate only $C_{y,k}$, while shared evidence about class $y$ and concept $k$ remains available elsewhere in the training data. We evaluate all methods on the same pairs and intervention manifests.

We measure representation quality using the same frozen-probe protocol as in \Secref{subsec:rep-quality}. After training each encoder, we freeze its representation and evaluate a linear probe on the unchanged test examples of the target pair; class-conditional concept Macro-F1 is the primary metric. We first average results across seeds within each pair, then obtain confidence intervals by bootstrapping the 24 target pairs (2,000 resamples). The \ccpair{} is therefore the experimental unit.



\subsubsection{Additional Results and Discussion}
\label{app:e4-results}

Table~\ref{tab:e4-full-retention} reports the complete \Secref{subsec:controlled-intervention} coverage-response results. Each intervention changes only the training evidence associated with a predefined target \ccpair{}. The corresponding class and concept remain represented elsewhere in the training set, isolating the ability to compose previously learned class and concept information from the simpler problem of recognizing an entirely unseen class or concept. We use the same target pairs for every method and retention level. We first average seeds within each pair, then estimate uncertainty by paired bootstrap over pairs, making the \ccpair{} the experimental unit.

\SemCovNet{} is stable across the full intervention range: representation quality $0.679$ at $100\%$ retention and $0.682$ at zero retention. Concept supervision also transfers substantial information to the unseen combination, explaining why its zero-coverage difference from \SemCovNet{} is positive but not statistically significant ($\Delta_0=+0.017$, $95\%$ CI $[-0.001,+0.036]$). In contrast, \SemCovNet{} significantly exceeds coverage reweighting and Semantic GroupDRO when there is zero evidence. This distinction demonstrates that both methods can emphasize poorly represented observed pairs, but neither creates new evidence when the exact pair is absent.

\begin{table}[t]
\centering
    \setlength{\tabcolsep}{4.5pt}
    \renewcommand{\arraystretch}{1.05}
    
    \caption{ \textbf{Full coverage-response intervention on CUB.}
     We report frozen-probe representation quality $q_{\mathrm{repr}}$ for each method as target-pair training evidence decreases from $100\%$ to zero. ``One'' retains a single positive example and therefore has no nominal retention fraction. Only the target pair's own evidence is intervened on; the corresponding class and concept remain observable through other training pairs. We report concept Macro-F1 as the mean over the target pairs with $95\%$ CIs. We average 5 seeds within each pair before bootstrapping. The final column reports the $\Delta_{100\%\!\rightarrow0\%}$, difference between 100\% and 0\% change. Smaller degradation indicates greater robustness to loss of pair-specific evidence; positive values indicate degradation, while values near zero indicate stability.}
    
    \label{tab:e4-full-retention}

    \resizebox{\linewidth}{!}
    {
    \begin{tabular}{lccccccc}
    \toprule
     Method & $100\%$ & $75\%$ & $50\%$ & $25\%$ & One & $0\%$ & $\Delta_{100\%\!\rightarrow0\%}$ \\
    \midrule
    Concept supervised 
    & 0.675 {\scriptsize [0.630, 0.717]} 
    & 0.666 {\scriptsize [0.619, 0.708]} 
    & 0.661 {\scriptsize [0.617, 0.703]} 
    & 0.665 {\scriptsize [0.619, 0.709]} 
    & 0.661 {\scriptsize [0.617, 0.704]} 
    & 0.665 {\scriptsize [0.619, 0.708]} 
    & 0.010  \\
    
    Coverage reweighting 
    & 0.650 {\scriptsize [0.608, 0.691]} 
    & 0.657 {\scriptsize [0.613, 0.698]} 
    & 0.659 {\scriptsize [0.618, 0.699]} 
    & 0.661 {\scriptsize [0.617, 0.702]} 
    & 0.661 {\scriptsize [0.620, 0.701]} 
    & 0.661 {\scriptsize [0.623, 0.697]} 
    & -0.012  \\
    
    Semantic GroupDRO 
    & 0.655 {\scriptsize [0.614, 0.696]} 
    & 0.658 {\scriptsize [0.617, 0.698]} 
    & 0.649 {\scriptsize [0.605, 0.689]} 
    & 0.659 {\scriptsize [0.618, 0.698]} 
    & 0.665 {\scriptsize [0.621, 0.707]} 
    & 0.660 {\scriptsize [0.617, 0.698]} 
    & -0.005 \\

    LoRA Concept 
    & 0.664 {\scriptsize [0.623, 0.704]} 
    & 0.658 {\scriptsize [0.618, 0.697]} 
    & 0.665 {\scriptsize [0.628, 0.702]} 
    & 0.669 {\scriptsize [0.628, 0.709]} 
    & 0.661 {\scriptsize [0.620, 0.703]} 
    & 0.670 {\scriptsize [0.629, 0.710]} 
    & -0.006 \\
    
    \SemCovNet{} (Ours)
    & \textbf{0.679 {\scriptsize [0.631, 0.724]}} 
    & \textbf{0.672 {\scriptsize [0.623, 0.717]} }
    & \textbf{0.677 {\scriptsize [0.629, 0.722]} }
    & \textbf{0.676 {\scriptsize [0.627, 0.721]} }
    & \textbf{0.678 {\scriptsize [0.631, 0.721]} }
    &\textbf{ 0.682 {\scriptsize [0.634, 0.726]} }
    & -0.003 \\
    
    \bottomrule
    \end{tabular}
    }
\end{table}

\begin{table}[t]
\centering
    \setlength{\tabcolsep}{6pt}
    \renewcommand{\arraystretch}{1.05}
\centering
\caption{
\textbf{Representation-sharing ablations under extreme semantic under-coverage.}
We evaluate structural variants in the one-example and zero-evidence conditions, where borrowing shared semantic structure matters most. We designed these variants as low-coverage mechanism ablations and therefore did not evaluate them over the full retention trajectory. We report results as means over the target \ccpairs{} with 95\% CIs. We average 5 seeds within each pair before bootstrapping.
}
\label{tab:e4-zero-ablation}
\begin{tabular}{lcc}
\toprule
Method & One example & Zero evidence \\
\midrule
No class sharing 
 & 0.678 {\scriptsize [0.627,0.724]}
 & 0.678 {\scriptsize [0.628,0.722]} \\

Shared prior only
 & 0.667 {\scriptsize [0.625,0.708]}
 & 0.667 {\scriptsize [0.624,0.709]} \\

Fixed specialization
 & 0.677 {\scriptsize [0.627,0.723]}
 & \textbf{0.684 {\scriptsize [0.635,0.728]}} \\

\CCSS{} (Ours)
 & \textbf{0.678 {\scriptsize [0.631,0.721]}}
 & 0.682 {\scriptsize [0.634,0.726]} \\
\bottomrule
\end{tabular}
\end{table}

Table~\ref{tab:e4-zero-ablation} results show that removing class sharing ($\mu_k$ only) changes the zero-coverage result by only $+0.004$ in favor of SemCovNet, with a confidence interval spanning zero. The coverage-controlled intervention alone does not statistically isolate the class-sharing representation's contribution at zero coverage. Similarly, several specialization variants remain competitive in this extreme condition. We therefore use this analysis to establish \textit{compositional recovery under missing pair evidence}, while separate component ablations assess which parts of the coverage-calibrated parameterization drive the overall gains.


\subsection{Semantic Coverage under Expert and Soft Clinical Supervision}
\label{app:medical-semantic-supervision}

This section provides the complete results underlying \Figref{fig:fig7-medical-generalization} in \Secref{sec:clinical-semantic-supervision}. We examine whether the coverage-aware semantic representation learning principle remains applicable across heterogeneous clinical semantic supervision. Derm7pt provides expert-defined categorical clinical concepts, whereas MILK10k provides continuous soft semantic evidence. We experiment with both datasets under identical settings for our proposed method; only the form of semantic supervision differs.

\textbf{Derm7pt.} 
\Figref{fig:fig7-medical-generalization}(a) reports frozen-probe class-conditioned concept Macro-F1 across Tail/Middle/Head coverage pairs. Table~\ref{tab:derm7pt-medical-generalization} provides the corresponding task and concept representation quality metrics across seeds. \SemCovNet{} improves Tail representation quality and substantially reduces the Head--Tail gap compared to standard concept supervision. Middle and Head performance are slightly lower, indicating that the gain is concentrated on low coverage clinical semantics rather than reflecting a uniform improvement across all pairs. The task-level results also improve in Macro-F1 and Balanced Accuracy, although \SemCovNet{} remains competitive with standard concept supervision on every concept metric.

Derm7pt further supports the association between coverage and representation quality through the frozen-probe analysis. However, its smaller sample size and diverse clinical appearances make the geometric dispersion trend less consistent across coverage groups, as shown in \Figref{fig:fig4b-coverage-representation}(c). We therefore interpret the Derm7pt result primarily through the coverage-stratified representation metrics rather than geometric compactness.

%

\begin{table*}[t]
\centering
    \setlength{\tabcolsep}{5pt}
    \renewcommand{\arraystretch}{1.05}
\caption{
\textbf{Expert clinical semantic supervision on Derm7pt.}
Task metrics evaluate diagnosis, while concept metrics evaluate the learned semantic representation. Tail AUROC and the $\Delta_{\mathrm{H-T}}$ Head--Tail gap use coverage strata defined from the training split; a smaller positive gap and smaller coverage slope $\beta_1$ indicate less dependence of representation quality on semantic coverage. Results are reported as mean $\pm$ standard deviation over five seeds.
}
\label{tab:derm7pt-medical-generalization}
\resizebox{\textwidth}{!}{
\begin{tabular}{lccc|ccc|c}
\toprule
\multirow{2}{*}{Method} & \multicolumn{3}{c|}{Task} & \multicolumn{3}{c|}{Concept} & 
\multirow{2}{*}{\makecell{Coverage\\Slope $\beta_1$ $\downarrow$}} \\
\cmidrule{2-4} \cmidrule{5-7}
~ & 
AUROC $\uparrow$ &
M-F1 $\uparrow$ &
BAcc $\uparrow$ &
 F1 $\uparrow$ &
Tail AUROC $\uparrow$ &
$\Delta_{\mathrm{H-T}}$ $\downarrow$ & ~ \\
\midrule

Concept supervised
& 0.848{\scriptsize$\pm$0.007}
& 0.738{\scriptsize$\pm$0.007}
& 0.717{\scriptsize$\pm$0.011}
& \textbf{0.485{\scriptsize$\pm$0.016}}
& 0.647{\scriptsize$\pm$0.049}
& 0.085{\scriptsize$\pm$0.049}
& 0.115{\scriptsize$\pm$0.030} \\

LoRA Concept
& 0.858{\scriptsize$\pm$0.005}
& 0.751{\scriptsize$\pm$0.012}
& 0.729{\scriptsize$\pm$0.014}
& 0.419{\scriptsize$\pm$0.196}
& 0.615{\scriptsize$\pm$0.021}
& 0.110{\scriptsize$\pm$0.021}
& 0.071{\scriptsize$\pm$0.020} \\


Semantic only
& \textbf{0.867{\scriptsize$\pm$0.006}}
& 0.761{\scriptsize$\pm$0.008}
& 0.745{\scriptsize$\pm$0.005}
& 0.404{\scriptsize$\pm$0.137}
& 0.683{\scriptsize$\pm$0.055}
& 0.038{\scriptsize$\pm$0.053}
& 0.101{\scriptsize$\pm$0.027} \\

Semantic + residual
& 0.855{\scriptsize$\pm$0.008}
& 0.762{\scriptsize$\pm$0.007}
& 0.743{\scriptsize$\pm$0.008}
& 0.490{\scriptsize$\pm$0.010}
& 0.655{\scriptsize$\pm$0.057}
& 0.063{\scriptsize$\pm$0.046}
& 0.092{\scriptsize$\pm$0.024} \\

Frozen \SemCovNet{}
& 0.852{\scriptsize$\pm$0.007}
& 0.756{\scriptsize$\pm$0.008}
& 0.736{\scriptsize$\pm$0.007}
& 0.365{\scriptsize$\pm$0.130}
& 0.673{\scriptsize$\pm$0.061}
& 0.054{\scriptsize$\pm$0.062}
& 0.093{\scriptsize$\pm$0.038} \\

\SemCovNet{} (Ours)
& 0.856{\scriptsize$\pm$0.007}
& \textbf{0.767{\scriptsize$\pm$0.011}}
& \textbf{0.759{\scriptsize$\pm$0.018}}
& 0.447{\scriptsize$\pm$0.023}
& \textbf{0.686{\scriptsize$\pm$0.020}}
& \textbf{0.025{\scriptsize$\pm$0.015}}
& \textbf{0.068{\scriptsize$\pm$0.010}} \\
\bottomrule
\end{tabular}}
\end{table*}

\begin{table*}[t]
\centering
    \setlength{\tabcolsep}{5pt}
    \renewcommand{\arraystretch}{1.05}
\caption{
\textbf{Soft semantic supervision on MILK10k.}
We perturb the continuous training evidence to make the semantic target less informative. Coverage strata are fixed from the unperturbed training evidence. Tail Concept F1 measures representation quality for under-covered semantics, Task AUROC measures clinical discrimination, and Soft Spearman measures agreement with the continuous semantic teacher. 
}
\label{tab:milk-soft}
\resizebox{0.70\textwidth}{!}
{
\begin{tabular}{c l ccc}
\toprule
$\eta$ &
Method &
Task AUROC $\uparrow$ &
Tail Concept F1 $\uparrow$ &
Soft Spearman $\uparrow$ \\
\midrule
\multirow{2}{*}{0.0} &
Concept supervised &
0.877{\scriptsize$\pm$0.017} &
0.785{\scriptsize$\pm$0.004} &
0.655{\scriptsize$\pm$0.114} \\

&
\textbf{\SemCovNet{}} &
0.887{\scriptsize$\pm$0.018} &
0.781{\scriptsize$\pm$0.021} &
0.637{\scriptsize$\pm$0.024} \\
\midrule

\multirow{2}{*}{0.1} &
Concept supervised &
0.879{\scriptsize$\pm$0.022} &
0.789{\scriptsize$\pm$0.001} &
0.643{\scriptsize$\pm$0.115} \\

&
\SemCovNet{} &
0.887{\scriptsize$\pm$0.019} &
0.786{\scriptsize$\pm$0.027} &
0.624{\scriptsize$\pm$0.022} \\
\midrule

\multirow{2}{*}{0.2} &
Concept supervised &
0.879{\scriptsize$\pm$0.021} &
0.790{\scriptsize$\pm$0.006} &
0.626{\scriptsize$\pm$0.111} \\

&
\textbf{\SemCovNet{}} &
0.888{\scriptsize$\pm$0.019} &
0.786{\scriptsize$\pm$0.025} &
0.604{\scriptsize$\pm$0.016} \\
\midrule

\multirow{2}{*}{0.3} &
Concept supervised &
0.879{\scriptsize$\pm$0.020} &
0.787{\scriptsize$\pm$0.006} &
0.608{\scriptsize$\pm$0.104} \\

&
\textbf{\SemCovNet{}} &
0.888{\scriptsize$\pm$0.018} &
0.788{\scriptsize$\pm$0.012} &
0.586{\scriptsize$\pm$0.014} \\
\midrule

\multirow{2}{*}{0.4} &
Concept supervised &
0.880{\scriptsize$\pm$0.021} &
0.783{\scriptsize$\pm$0.008} &
0.588{\scriptsize$\pm$0.098} \\

&
\textbf{\SemCovNet{}} &
0.888{\scriptsize$\pm$0.019} &
0.787{\scriptsize$\pm$0.013} &
0.572{\scriptsize$\pm$0.020} \\
\bottomrule
\end{tabular}}
\end{table*}

\textbf{MILK10k.} 
For MILK10k, we retain the original continuous semantic scores and perturb only the training evidence as 
\[ a'=(1-\eta)a+0.5\eta . \] 
Increasing $\eta$ makes the semantic evidence progressively less informative while preserving the continuous supervision interface. Tail/Middel/Head coverage pairs are defined once from the unperturbed training evidence and kept fixed across perturbation levels.

Table~\ref{tab:milk-soft} shows that \SemCovNet{} preserves both Tail semantic quality and task discrimination across this perturbation range. Soft-target reconstruction itself remains more competitive than concept supervision; this is expected because the objective is to learn coverage-aware visual representations rather than to reproduce the external soft concept score extractor as closely as possible.
Accordingly, this experiment is not intended to establish uniform robustness to semantic corruption. Instead, it shows that coverage-aware semantic representation learning remains applicable when semantic evidence is continuous and uncertain.

\subsection{Spurious-Correlation Robustness on Waterbirds}
\label{app:waterbirds-robustness}

WaterBirds provides a complementary test of whether coverage-aware semantic representation learning stays competitive under canonical spurious-correlation evaluation. WaterBirds defines four fixed groups from the target label (landbird/waterbird) and background (land/water). These groups, considered as pairs, are highly uneven in training coverage ($\text{Tail}=0.0502$, Middle $=0.50$, $\text{Head}=0.9498$): two pairs, waterbirds on land and landbirds on water, have coverage below $0.05$, although none has zero coverage. We therefore report standard group-robustness metrics.

\textbf{Group robustness.} 
Table~\ref{tab:waterbirds-group-robustness} reports average accuracy (Avg. Acc.), worst-group accuracy (WGA), and the Average--WGA gap. Dedicated group-robustness methods perform best on these metrics. SCER achieves the highest WGA ($0.923$) and the smallest Average--WGA gap ($0.027$), followed by Hierarchical-DRO and GroupDRO. This behavior is expected: these methods are designed to improve performance over the same fixed target$\times$background groups used by the WaterBirds evaluation.

\begin{table}[t]
    \centering
    \small
    \setlength{\tabcolsep}{5pt}
    \renewcommand{\arraystretch}{1.03}
    \caption{
        \textbf{Group robustness on WaterBirds.}
        WGA is the minimum accuracy across the four target$\times$background groups, and Avg--WGA measures the gap between average and worst-group accuracy.
    }
    \label{tab:waterbirds-group-robustness}
    \begin{tabular}{lccc}
        \toprule
        \textbf{Method}
        & \textbf{Avg. Acc. $\uparrow$}
        & \textbf{WGA $\uparrow$}
        & \textbf{Avg--WGA $\downarrow$} \\
        \midrule
        Concept supervised
        & 0.920{\scriptsize$\pm$0.016}
        & 0.816{\scriptsize$\pm$0.026}
        & 0.103{\scriptsize$\pm$0.017} \\

        JTT
        & 0.917{\scriptsize$\pm$0.010}
        & 0.822{\scriptsize$\pm$0.020}
        & 0.095{\scriptsize$\pm$0.015} \\

        MORE
        & 0.930{\scriptsize$\pm$0.003}
        & 0.819{\scriptsize$\pm$0.036}
        & 0.112{\scriptsize$\pm$0.039} \\

        IVQ-CBM
        & 0.910{\scriptsize$\pm$0.025}
        & 0.776{\scriptsize$\pm$0.021}
        & 0.134{\scriptsize$\pm$0.026} \\

        GroupDRO
        & 0.939{\scriptsize$\pm$0.003}
        & 0.882{\scriptsize$\pm$0.013}
        & 0.057{\scriptsize$\pm$0.013} \\

        Hierarchical-DRO
        & 0.944{\scriptsize$\pm$0.005}
        & 0.891{\scriptsize$\pm$0.011}
        & 0.052{\scriptsize$\pm$0.010} \\

        SCER
        & \textbf{0.950{\scriptsize$\pm$0.011}}
        & \textbf{0.923{\scriptsize$\pm$0.004}}
        & \textbf{0.027{\scriptsize$\pm$0.011}} \\

        \SemCovNet{} (Ours) 
        & 0.926{\scriptsize$\pm$0.008}
        & 0.830{\scriptsize$\pm$0.034}
        & 0.095{\scriptsize$\pm$0.034} \\
        \bottomrule
    \end{tabular}
\end{table}

\begin{table}[t]
    \centering
    \small
    \setlength{\tabcolsep}{4pt}
    \renewcommand{\arraystretch}{1.03}
    \caption{
        \textbf{Per-group accuracy on WaterBirds.}
        Groups correspond to the four combinations of bird class and background. Dedicated group-robustness methods particularly improve the conflicting landbird-on-water and waterbird-on-land groups.
    }
    \label{tab:waterbirds-per-group}
    \begin{tabular}{lcccc}
        \toprule
        \textbf{Method}
        & \textbf{Land/Land}
        & \textbf{Land/Water}
        & \textbf{Water/Land}
        & \textbf{Water/Water} \\
        \midrule

        Concept supervised
        & 0.997{\scriptsize$\pm$0.001} & 0.858{\scriptsize$\pm$0.036} & 0.816{\scriptsize$\pm$0.026} & 0.968{\scriptsize$\pm$0.005} \\

        JTT
        & 0.996{\scriptsize$\pm$0.001} & 0.861{\scriptsize$\pm$0.024} & 0.824{\scriptsize$\pm$0.026} & 0.973{\scriptsize$\pm$0.007} \\

        MORE
        & 0.997{\scriptsize$\pm$0.002} & 0.882{\scriptsize$\pm$0.021} & 0.823{\scriptsize$\pm$0.044} & 0.975{\scriptsize$\pm$0.009} \\

        IVQ-CBM
        & \textbf{0.998{\scriptsize$\pm$0.001}} & 0.871{\scriptsize$\pm$0.022} & 0.783{\scriptsize$\pm$0.020} & 0.956{\scriptsize$\pm$0.003} \\

        GroupDRO
        & 0.994{\scriptsize$\pm$0.002} & 0.894{\scriptsize$\pm$0.010} & 0.883{\scriptsize$\pm$0.014} & 0.955{\scriptsize$\pm$0.005} \\

        Hierarchical-DRO
        & 0.994{\scriptsize$\pm$0.001} & 0.905{\scriptsize$\pm$0.012} & 0.891{\scriptsize$\pm$0.011} & 0.956{\scriptsize$\pm$0.004} \\
        
        SCER
        & 0.977{\scriptsize$\pm$0.016} & \textbf{0.944{\scriptsize$\pm$0.010}} & \textbf{0.926{\scriptsize$\pm$0.003}} & 0.929{\scriptsize$\pm$0.014} \\

        \SemCovNet{} (Ours) 
        & 0.996{\scriptsize$\pm$0.001} & 0.868{\scriptsize$\pm$0.023} & 0.830{\scriptsize$\pm$0.034} & \textbf{0.975{\scriptsize$\pm$0.006}} \\

        \bottomrule
    \end{tabular}
\end{table}

\SemCovNet{} achieves $0.926$ average accuracy and $0.830$ WGA. It improves WGA over concept supervision ($0.816$) while reducing the Average--WGA gap from $0.103$ to $0.095$. However, it does not match methods that explicitly optimize the known spurious groups. We view this as complementary rather than contradictory to our main results: \SemCovNet{} is designed for uneven \cc{} coverage, not for direct worst-group optimization along a predefined nuisance axis.

\textbf{Per-group behavior.} 
The same pattern shows in the individual groups (Table~\ref{tab:waterbirds-per-group}). Concept supervised methods remain strong on aligned bird--background combinations but degrade on the conflicting combinations. \SemCovNet{} improves robustness relative to its matched non-group-aware controls, particularly without explicitly optimizing the WaterBirds group labels. In contrast, SCER, GroupDRO, and Hierarchical-DRO substantially improve the conflicting groups because their objectives directly target group-level robustness.

\textbf{Relation to SCI.} 
WaterBirds and SCI probe different forms of training imbalance. WGA evaluates minimum performance along one predefined group axis (bird class$\times$background) while the Average--WGA gap provides the disparity induced by that same partition. SCI instead analyzes how semantic coverage varies across the full set of \cc{} relations. Therefore, a method can perform strongly on WaterBirds by explicitly correcting the known background groups without addressing uneven coverage across a broader semantic vocabulary.

This distinction also explains why group-robustness methods perform better in this experiment. Their training objective and the WaterBirds evaluation use closely aligned group structure, whereas \SemCovNet{} does not optimize WGA directly. Nevertheless, \SemCovNet{} remains competitive in average accuracy and improves worst-group performance over concept supervised settings. These results suggest that coverage-aware semantic representation learning is compatible with canonical shortcut robustness while addressing a different, more fine-grained source of representation weakness.

\section{Ablation Study}
\label{app:ablation}

\subsection{What Component of Coverage-Aware Learning Matters?} 
\label{app:ablation-components}

 We first decompose \SemCovNet{} into concept supervision, shared semantic structure, pair-specific specialization, coverage calibration, and DINOv3 adaptation (Table~\ref{tab:ablation-components}). Each row changes one component compared to the initial configuration.

Concept supervision improves Tail Concept F1 from $0.528$ to $0.550$. Adding shared class and concept structure further increases it to $0.577$, while Mid and Head performance remains nearly unchanged. In contrast, fixed specialization provides no additional Tail gain, showing that simply adding pair-specific capacity is insufficient. Coverage-dependent specialization maintains or improves Tail performance across the calibrated variants. Finally, combining the full statistical CCSS mechanism with DINOv3 adaptation gives the strongest Tail result ($0.619\pm0.054$) and the largest reduction in the Head--Tail gap ($-0.069\pm0.055$).

Overall, the ablation suggests that the main benefit does not come from pair-specific specialization alone. Low coverage pairs benefit from shared semantic structure, while coverage calibration determines how much pair-specific evidence to use.

\subsection{Why CCSS Instead of a Simpler $\alpha$ Rule?} 
\label{app:ablation-alpha}

We test whether we can replace the specialization coefficient $\alpha_{y,k}$ with a simpler coverage-based rule (Table~\ref{tab:ablation-alpha}). We compare raw frequency, coverage alone, effective support, support with variance information, and the full CCSS estimator under the same training setting.

Using effective support already improves Tail Concept F1 over raw frequency ($0.625$ vs.\ $0.608$), showing that the amount of semantic evidence is more informative than a raw count alone. However, no particular statistic corresponds to the full CCSS rule. Full CCSS achieves the highest mean Tail performance ($0.634\pm0.057$), while Mid and Head performance remains stable. It also gives the largest Head--Tail gap reduction ($-0.087\pm0.060$).

These results support using a statistically calibrated specialization rule rather than a fixed coverage-based gate. CCSS jointly consider the available, reliable evidence when determining the extent to which a \ccpair{} should specialize.

\subsection{Shared Evidence for Low Coverage pairs}
\label{sec:shared-prior-ablation}

\begin{table}[!t]
    \centering
    \setlength{\tabcolsep}{3.5pt}
    \renewcommand{\arraystretch}{1.05}
    \caption{
    \textbf{Component decomposition of coverage-aware
    learning on CUB.}
    The ladder adds concept supervision, shared semantic structure,
    pair-specific specialization, coverage calibration, and 
    DINOv3 adaptation.
    }
    \label{tab:ablation-components}
    \resizebox{\linewidth}{!}{
    \begin{tabular}{lccclcc} 
        \toprule
        \textbf{Variant}
        & \textbf{Concept}
        & \textbf{Shared}
        & \textbf{Pair spec.}
        & \textbf{Calibration / adaptation}
        & \textbf{Tail F1$\uparrow$}
        & \textbf{$\Delta_{\mathrm{H-T}}$ $\downarrow$} \\
        \midrule
        ERM
        & -- & -- & -- & --
        & 0.528{\scriptsize$\pm$0.000}
        & 0.019{\scriptsize$\pm$0.000} \\

        Concept
        & \checkmark & -- & -- & --
        & 0.550{\scriptsize$\pm$0.039}
        & -0.005{\scriptsize$\pm$0.041} \\

        Shared
        & \checkmark & \checkmark & -- & --
        & 0.577{\scriptsize$\pm$0.037}
        & -0.032{\scriptsize$\pm$0.037} \\

        Fixed-spec
        & \checkmark & \checkmark & \checkmark & Fixed
        & 0.571{\scriptsize$\pm$0.029}
        & -0.020{\scriptsize$\pm$0.030} \\


        CCSS frozen
        & \checkmark & \checkmark & \checkmark & Statistical / frozen
        & 0.577{\scriptsize$\pm$0.039}
        & -0.028{\scriptsize$\pm$0.040} \\

        \SemCovNet{} (Ours) 
        & \checkmark & \checkmark & \checkmark & Statistical / LoRA-last3
        & \textbf{0.619{\scriptsize$\pm$0.054}}
        & \textbf{-0.069{\scriptsize$\pm$0.055}} \\
        \bottomrule
    \end{tabular}}
\end{table}

\begin{table}[!t]
    \centering
    \setlength{\tabcolsep}{4.5pt}
    \renewcommand{\arraystretch}{1.05}
    \caption{
    \textbf{Component ablation of coverage-aware learning on CUB.}
    Each row introduces one additional component, from concept supervision
    and shared semantic structure to pair-specific specialization, coverage
    calibration, and DINOv3 adaptation. We report results as mean $\pm$ standard deviation across five seeds. 
    }
    \label{tab:ablation-alpha}
    \begin{tabular}{lcc}
        \toprule
        \textbf{$\alpha$ rule}
        & \textbf{Tail F1$\uparrow$}
        & \textbf{$\Delta_{\mathrm{H-T}}\downarrow$} \\
        \midrule
        Frequency
        & 0.608{\scriptsize$\pm$0.057}
        & -0.062{\scriptsize$\pm$0.058} \\

        Coverage
        & 0.617{\scriptsize$\pm$0.042}
        & -0.072{\scriptsize$\pm$0.043} \\

        Effective support
        & 0.625{\scriptsize$\pm$0.028}
        & -0.079{\scriptsize$\pm$0.029} \\

        $n+\sigma^2$
        & 0.617{\scriptsize$\pm$0.042}
        & -0.069{\scriptsize$\pm$0.044} \\

        CCSS (Ours)
        & \textbf{0.634{\scriptsize$\pm$0.057}}
        & \textbf{-0.087{\scriptsize$\pm$0.060}} \\
        \bottomrule
    \end{tabular}
\end{table}

\begin{table}[!t]
\centering
    \setlength{\tabcolsep}{3.5pt}
    \renewcommand{\arraystretch}{1.05}
    \caption{\textbf{Shared-prior decomposition on CUB.} Effect of shared semantic structure on Tail, Middle, and Head representation quality.}
        
    \label{tab:ablation03-shared-prior}

    {
    \begin{tabular}{llcc}
    \toprule
    \textbf{Shared prior} & \textbf{Method} 
        & \textbf{Tail F1$\uparrow$}
        & \textbf{$\Delta_{\mathrm{H-T}}\downarrow$} \\
    \midrule
    Concept & $\tilde{\vmu}_k$ 
    & 0.579{\scriptsize$\pm$0.008}  
    & -0.032{\scriptsize$\pm$0.002} \\
    
    Class + concept & $\tilde{\vmu}_y+\tilde{\vmu}_k$ 
    &  0.598{\scriptsize$\pm$0.004}  
    & -0.047{\scriptsize$\pm$0.003}  \\
    
    Global + class + concept (Ours) & $\vmu_0+\tilde{\vmu}_y+\tilde{\vmu}_k$ 
    & \textbf{0.608{\scriptsize$\pm$0.007}}  
    & \textbf{-0.057{\scriptsize$\pm$0.005}}  \\
    \bottomrule
    \end{tabular}
    }
\end{table}

Table~\ref{tab:ablation03-shared-prior} isolates the shared structure used by coverage-aware learning. Tail representation quality improves progressively from concept-level evidence to class-shared structure to combining global, class, and concept evidence sharing ($0.579\rightarrow0.598\rightarrow0.608$, respectively); Mid and Head performance remains consistent. 
These results show that low coverage \ccpairs{} should not be learned independently from limited paired evidence. They benefit from progressively more general semantic structure (concept, class, and global) that provides a more stable sharing prior until sufficient pair-specific evidence supports specialization.
This supports structured semantic sharing across class and concept as the prior for coverage-aware specialization.

\begin{table}[!t]
    \centering
    \setlength{\tabcolsep}{4.5pt}
    \renewcommand{\arraystretch}{1.05}
    \caption{
        \textbf{Sensitivity to $\lambda_{\mathrm{CCSS}}$ on CUB.}
        We vary the CCSS weight around the main setting $\lambda_{\mathrm{CCSS}}^{*}=0.25$. 
        $\Delta$ reports the paired change in Tail Macro-F1 relative to $\lambda_{\mathrm{CCSS}}^{*}$ with 95\% CIs.
    }
    \label{tab:lambda-ccss-sensitivity}
    \begin{tabular}{c c c c c}
        \toprule
        \textbf{$\lambda/\lambda^{*}$}
        & \textbf{$\lambda_{\mathrm{CCSS}}$}
        & \textbf{Tail F1 $\uparrow$}
        & \textbf{H--T Gap $\downarrow$}
        & \textbf{$\Delta$ vs.\ $\lambda^{*}$ [95\% CI]} \\
        \midrule
        $0.25\times$ & 0.0625 & 0.580{\scriptsize$\pm$0.056} & -0.033{\scriptsize$\pm$0.056}
        & -0.006 {\scriptsize[-0.020, +0.006]} \\
        $0.5\times$  & 0.1250 & 0.606{\scriptsize$\pm$0.056} & -0.058{\scriptsize$\pm$0.055}
        & -0.002 {\scriptsize[-0.011, +0.007]} \\
        $1\times$    & 0.2500 & 0.599{\scriptsize$\pm$0.071} & -0.047{\scriptsize$\pm$0.072}
        & -- \\
        $2\times$    & 0.5000 & 0.617{\scriptsize$\pm$0.068} & -0.065{\scriptsize$\pm$0.069}
        & +0.008 {\scriptsize[-0.001, +0.015]} \\
        
        $4\times$    & 1.0000 & 0.608{\scriptsize$\pm$0.057} & -0.056{\scriptsize$\pm$0.056}
        & +0.008 {\scriptsize[-0.007, +0.021]} \\
        \bottomrule
    \end{tabular}
\end{table}

\subsection{Sensitivity to the CCSS Weight} 
\label{app:lambda-ccss-sensitivity}

We investigate sensitivity to the CCSS loss weight $\lambda_{\mathrm{CCSS}}$ on CUB by varying it from $0.25\times$ to $4\times$ the value used in the main experiments ($\lambda_{\mathrm{CCSS}}^{*}=0.25$). Across this range, Tail concept performance remains stable, with Macro-F1 between $0.580$ and $0.617$, while Head performance changes little (Table~\ref{tab:lambda-ccss-sensitivity}). The corresponding mean specialization coefficient $\alpha$ also remains stable.

Although $2\times\lambda_{\mathrm{CCSS}}^{*}$ gives the highest numerical Tail score, none of the tested settings differs significantly from the chosen $\lambda_{\mathrm{CCSS}}^{*}$ in the paired analysis over the 24 evaluation pairs. This shows that the improvements in coverage-aware representation are consistent and do not rely on precise tuning of $\lambda_{\mathrm{CCSS}}$.

\end{document}